\documentclass[final,5p,times,twocolumn,sort&compress]{elsarticle}

\AtBeginDocument{%
  \def\thefnote{\myfnsymbol{fnote}}}
\makeatletter
\def\myfnsymbol#1{\expandafter\@myfnsymbol\csname c@#1\endcsname}
\def\@myfnsymbol#1{\ifcase #1\or $\#$\or $\#\#$\else \@ctrerr\fi}
\def\fntext[#1]#2{\g@addto@macro\@fnotes{%
   \refstepcounter{fnote}\elsLabel{#1}%
   \def\thefootnote{\thefnote}% <-- corrected
   \global\setcounter{footnote}{\c@fnote}%
   \footnotetext{#2}}}
\makeatother

\usepackage{graphicx} \graphicspath{{figures/}}
\usepackage{amsmath,amssymb,mathbbol}
\usepackage{acronym}
\usepackage{enumitem}
\usepackage[pagebackref,breaklinks,colorlinks]{hyperref}
\usepackage{xspace,setspace}
\usepackage[skip=3pt,font=small]{subcaption}
\usepackage[skip=3pt,font=small]{caption}
\usepackage[dvipsnames,svgnames,x11names]{xcolor}
\usepackage[capitalise,nameinlink]{cleveref}
\usepackage{booktabs,tabularx,colortbl,multirow,array,makecell}
\usepackage{overpic,wrapfig}
\usepackage{stfloats}
\makeatletter
\def\ps@pprintTitle{%
 \let\@oddhead\@empty
 \let\@evenhead\@empty
 \def\@oddfoot{}%
 \let\@evenfoot\@oddfoot}
\makeatother

\makeatletter
\renewcommand{\paragraph}{%
  \@startsection{paragraph}{4}%
  {\z@}{0ex \@plus 0ex \@minus 0ex}{-1em}%
  {\hskip\parindent\normalfont\normalsize\itshape}%
}
\makeatother

\journal{Engineering}

\hfuzz=\maxdimen
\vfuzz=\maxdimen

\makeatletter
\DeclareRobustCommand\onedot{\futurelet\@let@token\@onedot}
\def\@onedot{\ifx\@let@token.\else.\null\fi\xspace}
\def\eg{\textit{e.g}\onedot} 

\def\ie{\textit{i.e}\onedot}

\def\etal{\textit{et al}\onedot}
\makeatother

\begin{document}

\begin{frontmatter}

\title{A Reconfigurable Data Glove for Reconstructing Physical and Virtual Grasps}

\author[nkl]{Hangxin Liu%
\texorpdfstring{\corref{ca}}{}%
\texorpdfstring{\fnref{eq}}{}}
\ead{liuhx@bigai.ai}
\author[nkl,ucla]{Zeyu Zhang%
\texorpdfstring{\fnref{eq}}{}}
\author[nkl,ucla]{Ziyuan Jiao%
\texorpdfstring{\fnref{eq}}{}}
\author[nkl]{Zhenliang Zhang}
\author[math]{\\Minchen Li}
\author[math]{Chenfanfu Jiang}
\author[pku]{Yixin Zhu%
\texorpdfstring{\corref{ca}}{}}
\ead{yixin.zhu@pku.edu.cn}
\author[nkl,pku,thu]{Song-Chun~Zhu}

\cortext[ca]{Corresponding authors.}
\fntext[eq]{These authors contributed equally to this work.}

\address[nkl]{National Key Laboratory of General Artificial Intelligence, Beijing Institute for General Artificial Intelligence (BIGAI)}
\address[ucla]{Center for Vision, Cognition, Learning, and Autonomy (VCLA), Department of Statistics, UCLA}
\address[math]{Multi-Physics Lagrangian-Eulerian Simulations Lab, Department of Mathematics, UCLA}
\address[pku]{Institute for Artificial Intelligence, Peking University}
\address[thu]{Department of Automation, Tsinghua University}

\begin{abstract}
In this work, we present a reconfigurable data glove design to capture different modes of human hand-object interactions, which are critical in training embodied artificial intelligence (AI) agents for fine manipulation tasks. To achieve various downstream tasks with distinct features, our reconfigurable data glove operates in three modes sharing a unified backbone design that reconstructs hand gestures in real time. In the \textit{tactile-sensing} mode, the glove system aggregates manipulation force via customized force sensors made from a soft and thin piezoresistive material; this design minimizes interference during complex hand movements. The \textit{virtual reality (VR)} mode enables real-time interaction in a physically plausible fashion: A caging-based approach is devised to determine stable grasps by detecting collision events. Leveraging a state-of-the-art finite element method (FEM), the \textit{simulation mode} collects data on fine-grained 4D manipulation events comprising hand and object motions in 3D space and how the object's physical properties (\eg, stress and energy) change in accordance with manipulation over time. Notably, the glove system presented here is the first to use high-fidelity simulation to investigate the unobservable physical and causal factors behind manipulation actions. In a series of experiments, we characterize our data glove in terms of individual sensors and the overall system. More specifically, we evaluate the system's three modes by (i) recording hand gestures and associated forces, (ii) improving manipulation fluency in VR, and (iii) producing realistic simulation effects of various tool uses, respectively. Based on these three modes, our reconfigurable data glove collects and reconstructs fine-grained human grasp data in both physical and virtual environments, thereby opening up new avenues for the learning of manipulation skills for embodied AI agents.
\end{abstract}

\end{frontmatter}

% \linenumbers
\setcounter{footnote}{0}

\section{Challenges in learning manipulation}

Manipulation and grasping are among the most fundamental topics in robotics. This classic field has been rejuvenated by the recent boom in embodied AI, wherein an agent (\eg, a robot) is tasked to learn by interacting with its environment. Since then, learning-based methods have been widely applied and have elevated robots' manipulation competence. Often, robots either train on data directly obtained from sensors (\eg, object grasping from a cluster~\cite{pinto2016supersizing,mahler2019learning}, pick-and-place~\cite{zeng2018robotic}, object handover~\cite{cini2019choice}, or door opening~\cite{yahya2017collective}) or learn from human demonstrations (\eg, motor motion~\cite{schaal2003computational,maeda2016acquiring}, affordance~\cite{nguyen2016detecting,kokic2017affordance}, task structure~\cite{mohseni2015interactive,xiong2016robot,liu2019mirroring}, or reward functions~\cite{abbeel2004apprenticeship,prieur2012modeling,ibarz2018reward}). 

Learning meaningful manipulation has a unique prerequisite: It must incorporate fine-grained physics to convey an understanding of the complex process that occurs during the interaction. Although we have witnessed the solid advancement of certain embodied AI tasks (\eg, visual-language navigation), these successes are primarily attributed to the readily available plain images and their annotations (pixels, segments, or bounding boxes) that are directly extracted from the existing training platforms~\cite{xie2019vrgym,li2021igibson,szot2021habitat}, while physics information during the interactions is still lacking. Similarly, although modern vision-based sensors and motion-capture systems can collect precise trajectory information, neither can precisely estimate physical properties during interactions. Existing software and hardware systems are insufficient for learning sophisticated manipulation skills for the following three reasons: 

\begin{figure*}[t!]
	\centering
	\begin{overpic}
        [width=\linewidth]{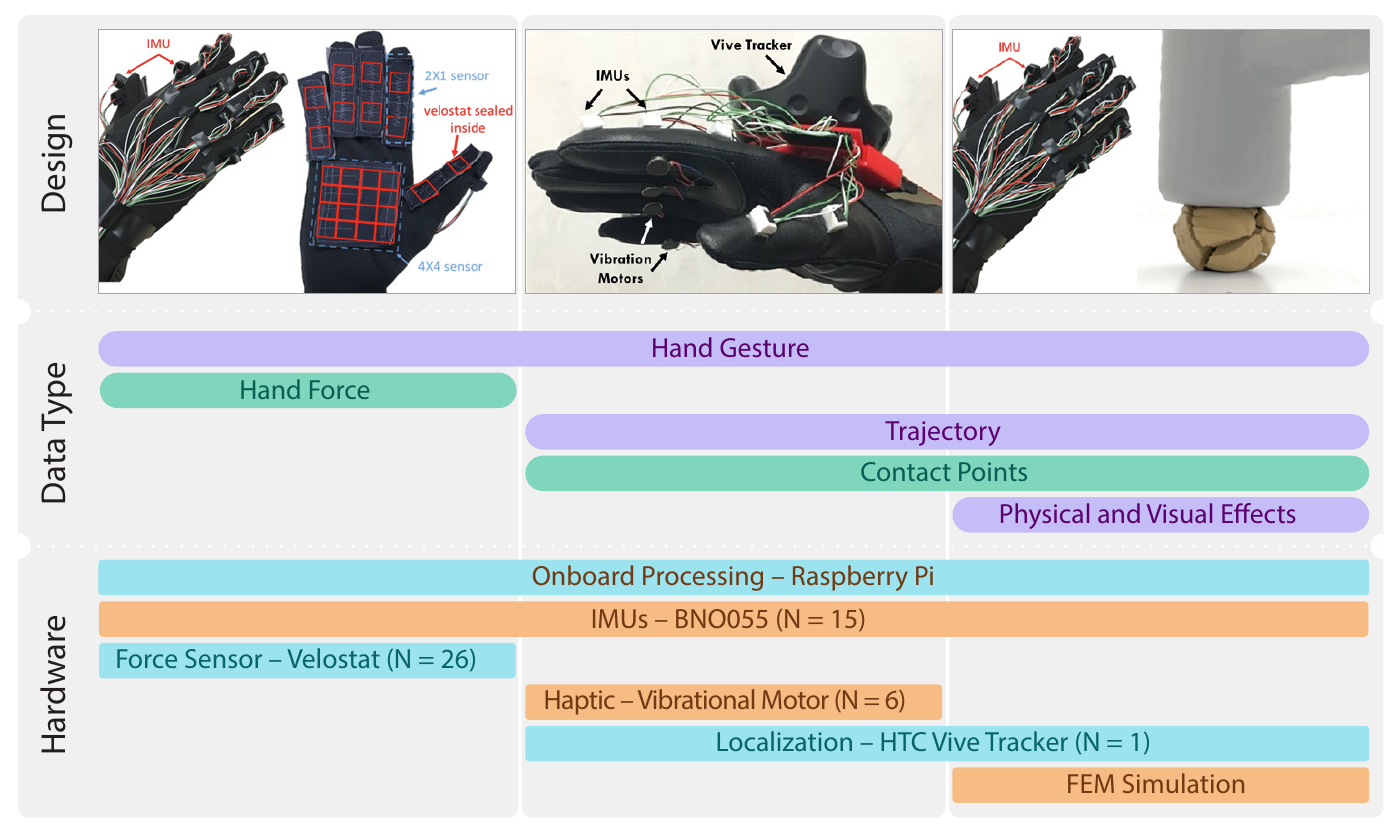}
	    \put(10,36.6){\color{black} \textbf{(a) Tactile-sensing mode}~\cite{liu2017glove}}
	    \put(45,36.6){\color{black} \textbf{(b) VR mode}~\cite{liu2019high}}
	    \put(75,36.6){\color{black} \textbf{(c) Simulation mode}}
	\end{overpic}
	\caption{\textbf{Overview of our reconfigurable data glove in three operating modes, which share a unified backbone design of an IMU network that captures the hand gesture.} (a) The tactile-sensing mode records the force exerted by the hand during manipulation. (b) The VR mode supports stable grasping of virtual objects in VR applications and provides haptic feedback via vibration motors. Contact configurations are conveniently logged. (c) The simulation mode incorporates state-of-the-art FEM simulation~\cite{li2020incremental} to augment the grasp data with fine-grained changes in the object's properties.}
	\label{fig:summary}
\end{figure*}

First, understanding fine-grained manipulation or human-object interactions requires a joint understanding of both hand gesture\footnote{In this article, the phrase ``hand gesture'' is used to refer to the collective movement of the fingers and palm, whereas ``hand pose'' is used to refer to the position and orientation of the wrist.} and force~\cite{liu2017glove}; distinguishing certain actions purely based on the hand gesture is challenging, if not impossible. For example, in the task of opening a medicine bottle that requires either pushing or squeezing the lid to unlock the childproof mechanism, it is insufficient to differentiate the opening actions by visual information alone, because the pushing and squeezing actions are visually similar (or even identical) to each other~\cite{edmonds2019tale}. Reconstructing hand gestures or trajectories alone has already been shown to be challenging, as severe hand-object occlusion hinders the data collection reliability. To tackle this problem, we introduce a \textit{tactile-sensing glove} to jointly capture hand gestures through a network of inertial measurement units (IMUs) and force exerted by the hand using six customized force sensors during manipulation. The force sensors are constructed from Velostat--a piezoresistive fabric with changing resistance under pressures, which is soft and thin to allow natural hand motions. Together, the force sensors provide a holistic view of manipulation events. A preliminary version of this system has been presented in the work of Liu \etal~\cite{liu2017glove}; interested readers can refer to the Appendix for details.

Second, contact points between hand and object play a significant role in understanding why and how a specific grasp is chosen. Such information is traditionally challenging to obtain (\eg, through thermal imaging~\cite{brahmbhatt2019contactdb}). To address this challenge, we devise a \textit{VR glove} and leverage VR platforms to obtain contact points. This design incorporates a caging-based approach to determine a stable grasp of a virtual object based on the collision geometry between fingers and the object. The collisions trigger a network of vibration motors on the glove to provide haptic feedback. The VR glove jointly collects trajectory and contact information that is otherwise difficult to obtain physically. A preliminary version of this system has been presented in the work of Liu \etal~\cite{liu2019high}; interested readers can refer to the Appendix for details.

Third, much attention has been paid to collecting hand information during fine manipulation but not to the object being manipulated or its effects caused by actions. This deficiency prohibits the use of collected data for studying complex manipulation events. For example, consider a tool-use scenario. A manipulation event cannot be comprehensively understood without capturing the \textit{interplay} between the human hand, the tool being manipulated, and the action effects. As such, this perspective demands a solution beyond the classic hand-centric view in developing data gloves. Furthermore, since the effects caused by the manipulation actions are traditionally difficult to capture, they are often treated as a task of recognizing discrete, symbolic states or attributes in computer vision~\cite{duan2012discovering,liu2017jointly,nagarajan2018attributes}, losing their intrinsic continuous nature. To overcome these limits of traditional data gloves, we propose to integrate a physics-based simulation using the state-of-the-art FEM~\cite{li2020incremental} to model object fluents--the time-varying states in the event~\cite{newton1736method}--and other physical properties involved, such as contact forces and the stress within the object. This \textit{glove with simulation} captures a human manipulation action and analyzes it in four-dimensional (4D) space by including: (i) the contact and geometric information of the hand gesture and the object in three-dimensional (3D) space, and (ii) the transition and coherence between the object's fluent changes and the manipulation events over time. To the best of our knowledge, this is the first time such 4D data offering a holistic view of manipulation events is used in this field, and its use will open up new avenues for studying manipulations and grasping.

Sharing a unified backbone design that reconstructs hand gestures in real-time, the proposed data glove can be easily reconfigured to (i) capture force exerted by hand using piezoresistive material, (ii) record contact information by grasping stably in VR, or (iii) reconstruct both visual and physical effects during the manipulation by integrating physics-based simulation. Our system extends the long history of developing data gloves~\cite{dipietro2008survey} and endows embodied AI agents with a deeper understanding of hand-object interactions.

This paper makes three contributions compared with prior work~\cite{liu2017glove,liu2019high}. First, we introduce the concept of a reconfigurable glove-based system. The three operating modes tackle a broader range of downstream tasks with distinct features. This extension does not sacrifice the easy-to-replicate nature, as different modes share a unified backbone design. Second, a state-of-the-art FEM-based physical simulation is integrated to augment the grasp data with simulated action effects, thereby providing new opportunities for studying hand-object interactions and complex manipulation events. Third, we demonstrate that the data collected by our glove-based system--either virtually or physically--is effective for learning in a series of case studies.

\subsection{Related work}

\paragraph{Hand gesture sensing}

Recording finger joints' movements is the core of hand gesture sensing. Various types of hardware have been adopted to acquire hand gestures. Although curvature/flex sensors~\cite{kramer2011soft,kamel2008glove}, liquid metal~\cite{oh2021liquid}, a stretchable strain sensor~\cite{wang2020gesture}, and triboelectric material~\cite{wen2020machine} are among proven approaches, these can only measure unidirectional bending angles. Hence, they are less efficient for recording a hand's metacarpophalangeal (MCP) joints with two degrees of freedom (DoFs) for finger abduction and adduction. In addition, by wrapping around bending finger joints, these instruments sacrifice natural hand movements due to their large footprint and rigidness. In comparison, IMUs can measure one phalanx's 6-DoF pose, interfere less with joint motions, and perform more consistently over an extended period of time. As a result, adopting IMUs in data gloves has prevailed in modern design, including IMUs channeled by a Zigbee network~\cite{taylor2013forward}, a circuit board with a 6-DoF accelerometer/gyroscope and a 3-DoF magnetometer placed on each of the 15 phalanxes~\cite{kortier2014assessment}, and a population of IMUs connected through flexible cables~\cite{hu2020flexible}. Often, the raw sensory information requires further filtering~\cite{santaera2015low} and estimation~\cite{kortier2014assessment,ligorio2013extended,kortier2015hand}.

\paragraph{Force sensing}

Sensing the forces exerted by a hand during manipulation has attracted growing research attention and requires a more integrated glove-based system. Here, we highlight some signature designs. An elastomer sensor with embedded liquid-metal material~\cite{hammond2014toward} was able to sense force across a large area (\eg, the palm) and estimate joint movements by measuring skin strain. FlexiForce sensors can acquire hand forces~\cite{gu2015fine}, while an optical-based motion-capture system tracks hand gestures. Forces and gestures can also be estimated using 9-DoF IMUs without additional hardware~\cite{mohammadi2016fingertip}, although the force estimation is crude. Other notable designs involve specialized hardware, including force-sensitive resistors~\cite{lin2019novel} and a specific tactile sensor for fingertips~\cite{battaglia2016thimblesense}. Recently, soft films made from piezoresistive materials whose resistance changes under pressing forces (\eg, Velostat) have become increasingly popular in robotic applications; this type of material permits force sensing without constraining the robots' or human hand's motions~\cite{low2015pressure,pugach2016touch,muller2015smart,jeong2011finger}.

\subsection{Overview: Three modes of the reconfigurable data glove}

To tackle the aforementioned challenges and fill in the gap in the literature, we devised a reconfigurable data glove that is capable of operating in three modes for various downstream tasks with distinct features and goals.

\paragraph{Tactile-sensing mode}

We start with a glove design using an IMU configuration~\cite{kortier2014assessment} to reconstruct hand gestures. Our system's software and hardware designs are publicly available for easy replication. A customized force sensor made from Velostat--a soft fabric whose resistance changes under different pressures--is adopted to acquire the force distributions over large areas of the hand without constraining natural hand motions. \cref{fig:summary}a~\cite{liu2017glove,liu2019high,li2020incremental} summarizes this tactile-sensing glove design.

\paragraph{VR mode}

By reconstructing virtual grasps in VR, this mode provides supplementary contact information (\eg, contact points on an object) during manipulation actions. In contrast to the dominating symbolic grasp methods that directly attach the virtual object to the virtual hand when a grasp event is triggered~\cite{boulic1996multi}, our glove-based system enables a natural and realistic grasp experience with a fine-grained hand gesture reconstruction and force estimated at specific contact points; a symbolic grasp would cause finger penetrations or non-contacting (\eg, see examples in \cref{fig:vr:comparison_touch}), since the attachments between the hand and object are predefined. Although collecting grasp-related data in VR is more convenient and economical than other specialized data-acquisition pipelines, the lack of direct contact between the hand and physical objects inevitably leads to less natural interactions. Thus, providing haptic feedback is critical to compensate for this drawback. We use vibration motors to provide generic haptic feedback to each finger, thereby increasing the realism of grasping in VR. \cref{fig:summary}b~\cite{liu2017glove,liu2019high,li2020incremental} summarizes the VR glove design.

\paragraph{Simulation mode}

Physics-based simulations emulate a system's precise changes over time, thus opening up new directions for robot learning\cite{choi2021use}, including learning robot navigation~\cite{xie2019vrgym}, bridging human and robot embodiments in learning from demonstration~\cite{liu2019mirroring}, soft robot locomotion~\cite{hu2019chainqueen}, liquid pouring~\cite{kennedy2019autonomous}, and robot cutting~\cite{heiden2021disect}. In a similar vein, simulating how an object's fluent changes as the result of a given manipulation action provides a new perspective on hand-object interactions. In this article, we adopt a state-of-the-art FEM simulator~\cite{li2020incremental} to emulate the causes and effects of manipulation events. As shown in \cref{fig:summary}c~\cite{liu2017glove,liu2019high,li2020incremental}, by integrating physical data collected by the data glove with simulated effects, our system reconstructs a new type of 4D manipulation data with high-fidelity visual and physical properties on a large scale. We believe that this new type of data can significantly impact how manipulation datasets are collected in the future and can assist in a wide range of manipulation tasks in robot learning.

\subsection{Structure of this article}

The remainder of this article is organized as follows. We start with a unified design for hand gesture sensing in\cref{sec:gesture}. With different goals, the tactile-sensing mode~\cite{liu2017glove} and the VR mode~\cite{liu2019high} are presented in \cref{sec:tactile} and \cref{sec:vr}, respectively. A new state-of-the-art, physics-based simulation using FEM~\cite{wolper2019cd} is integrated in \cref{sec:sim} to collect 4D manipulation data, which is the very first in the field to achieve such high fidelity, to the best of our knowledge. We evaluate our system in three modes in \cref{sec:application} and conclude the paper in \cref{sec:conclusion}.

\section{A unified backbone design for gesture sensing}\label{sec:gesture}

This section introduces the IMU setup for capturing hand gestures in \cref{sec:gesture:reconst}. As this setup is shared among all three modes of the proposed reconfigurable data glove, we further evaluate the IMU performance in \cref{sec:gesture:imu_eval}.

\subsection{Hand gesture reconstruction}\label{sec:gesture:reconst}

\paragraph{IMU specification}

Fifteen Bosch BNO055 9-DoF IMUs are deployed for hand gesture sensing. One IMU is mounted to the palm, two IMUs to the thumb's distal and intermediate phalanges, and the remaining 12 are placed on the phalanxes of the other four fingers. Each IMU includes a 16-bit triaxial gyroscope, a 12-bit triaxial accelerometer, and a triaxial geomagnetometer. This IMU is integrated with a built-in proprietary sensor fusion algorithm running on a 32-bit microcontroller, yielding each phalanx's pose in terms of a quaternion. The geomagnetometer acquires an IMU's reference frame to the Earth's magnetic field, supporting the pose calibration protocol (introduced later). The small footprint of the BNO055 ($5~cm \times 4.5~cm$) allows easy attachment to the glove and minimizes interference with natural hand motions. A pair of TCA9548A $I^2C$ multiplexers is used for networking the 15 IMUs and connecting them to the $I^2C$ bus interfaces on a Raspberry Pi 2 Model B board (henceforth RPi for brevity); RPi acts as the master controller for the entire glove system.

\paragraph{Hand forward kinematics}

A human hand has about 20 DoFs: both the proximal interphalangeal (PIP) joint and the distal interphalangeal (DIP) joint have one DoF, whereas an MCP joint has two. Based on this anatomical structure, we model each finger by a 4-DoF kinematic chain whose base frame is the palm and the end-effector frame is the distal phalanx. The thumb is modeled as a 3-DoF kinematic chain consisting of a DIP joint and an MCP joint.

\begin{figure}[t!]
	\centering
	\includegraphics[width=0.7\linewidth]{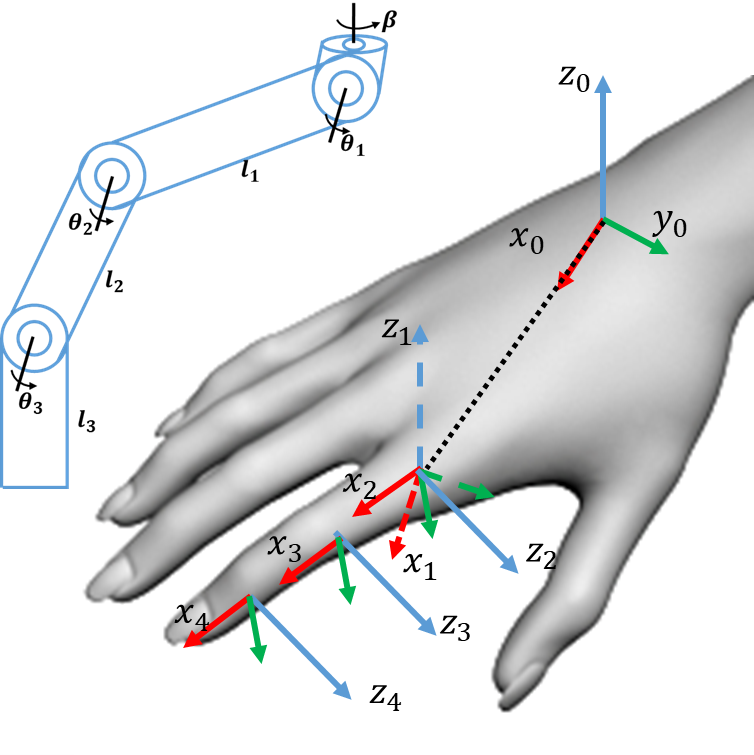}
	\caption{\textbf{The kinematic chain of the index finger with coordinate frames attached.} Reproduced from Ref.~\cite{liu2017glove} with permission.}
	\label{fig:gesture:hand_kin}
\end{figure}

After obtaining a joint's rotational angle using two consecutive IMUs, the position and orientation of each phalanx can be computed by forward kinematics. \cref{fig:gesture:hand_kin}~\cite{liu2017glove} shows an example of the index finger's kinematic chain and the attached frame. Frame 1 is assigned to the palm, and Frames 2, 3, and 4 are assigned to the proximal, middle, and distal phalanx, respectively. The proximal, middle, and distal phalanx lengths are respectively denoted by $l_1$, $l_2$, and $l_3$, The flexion and extension angles of the MCP, PIP, and DIP joints are denoted as $\theta_1$, $\theta_2$, and $\theta_3$, respectively. In addition, the MCP joint has an abduction and adduction angle denoted as $\beta$. $d_x$ and $d_y$ are the offsets in the $x$ and $y$ directions between the palm's center and the MCP joint. \cref{table:gesture:DH} derives the Denavit-Hartenberg (D-H) parameters for each reference frame, wherein a general homogeneous transformation matrix $\mathbf{T}$ from frames $i-1$ to $i$ can be given by the following:
\begin{equation}
    ^{i-1}_{i}\mathbf{T} =
    \begin{bmatrix}
        c\theta_i & -s\theta_i & 0 & a_{i-1} \\
        s\theta_{i}c\alpha_{i-1} & c\theta_{i}c\alpha_{i-1} & -s\alpha_{i-1} & -s\alpha_{i-1}d_i \\
        s\theta_{i}s\alpha_{i-1} & c\theta_{i}s\alpha_{i-1} & c\alpha_{i-1} & c\alpha_{i-1}d_i \\
        0 & 0 & 0 & 1
    \end{bmatrix},
\end{equation}
where $c\theta_i$ and $s\theta_i$ denote $\cos{(\theta_i)}$ and $\sin{(\theta_i)}$, respectively.

\begin{table}[ht!]
    \centering
    \small
    \caption{\textbf{Denavit-Hartenberg parameters of a finger.}}
    \label{table:gesture:DH}
    \begin{tabular}{ccccc}
        \toprule
        Link ID & $\alpha_{i-1}$ & $a_{i-1}$ & $\theta_{i}$ & $d_{i}$\\
        \midrule
        1 & 0 & 0 & $\beta$ & 0\\
        2 & ${\pi}/{2}$ & $l_1$ & $\theta_1$ & 0\\
        3 & 0 & $l_2$ & $\theta_2$ & 0\\
        4 & 0 & $l_3$ & $\theta_3$ & 0\\
        \bottomrule
    \end{tabular}
\end{table}

\cref{table:gesture:trans} lists the homogeneous transformation matrices of each phalanx, which can be used to express each phalanx's pose in the palm's reference frame in the cartesian space. The forward kinematics model keeps better track of the sensed hand gesture by reducing the inconsistency due to IMU fabrication error and anatomical variations among the users' hands.

\begin{table}[ht!]
    \centering
    \small
    \caption{\textbf{Concatenation of transformation matrices.}}
    \label{table:gesture:trans}
    \begin{tabular}{cc}
        \toprule
        Phalanx & Transformation \\
        \midrule
        Proximal & $^{0}_{1}T ^{1}_{2}T$ \\
        Middle/Distal for thumb & $^{0}_{1}T ^{1}_{2}T ^{2}_{3}T$\\
        Distal  & $^{0}_{1}T ^{1}_{2}T ^{2}_{3}T ^{3}_{4}T$\\
        \midrule
    \end{tabular}
\end{table}

\paragraph{Joint limits}

We adopt a set of commonly used inequality constraints~\cite{lin2000modeling} to limit the motion ranges of the finger joints, thereby eliminating unnatural hand gestures due to sensor noise:

\begin{equation}
    \begin{aligned}
        \text{MCP joint} &: 
        \begin{cases}
        0^{\circ} \le \theta_{1} \le 90^{\circ} \\
        -15^{\circ} \le \beta \le 15^{\circ} \\
        \end{cases}
        \\
        \text{PIP joint} &:
        0^{\circ} \le \theta_{2} \le 110^{\circ}  
        \\
        \text{DIP joint} &: 
        0^{\circ} \le \theta_{3} \le 90^{\circ} 
    \end{aligned}
    \label{eq:joint_limit_F}
\end{equation}

\paragraph{Pose calibration}

\begin{figure}[t!]
    \centering
    \includegraphics[width=\linewidth]{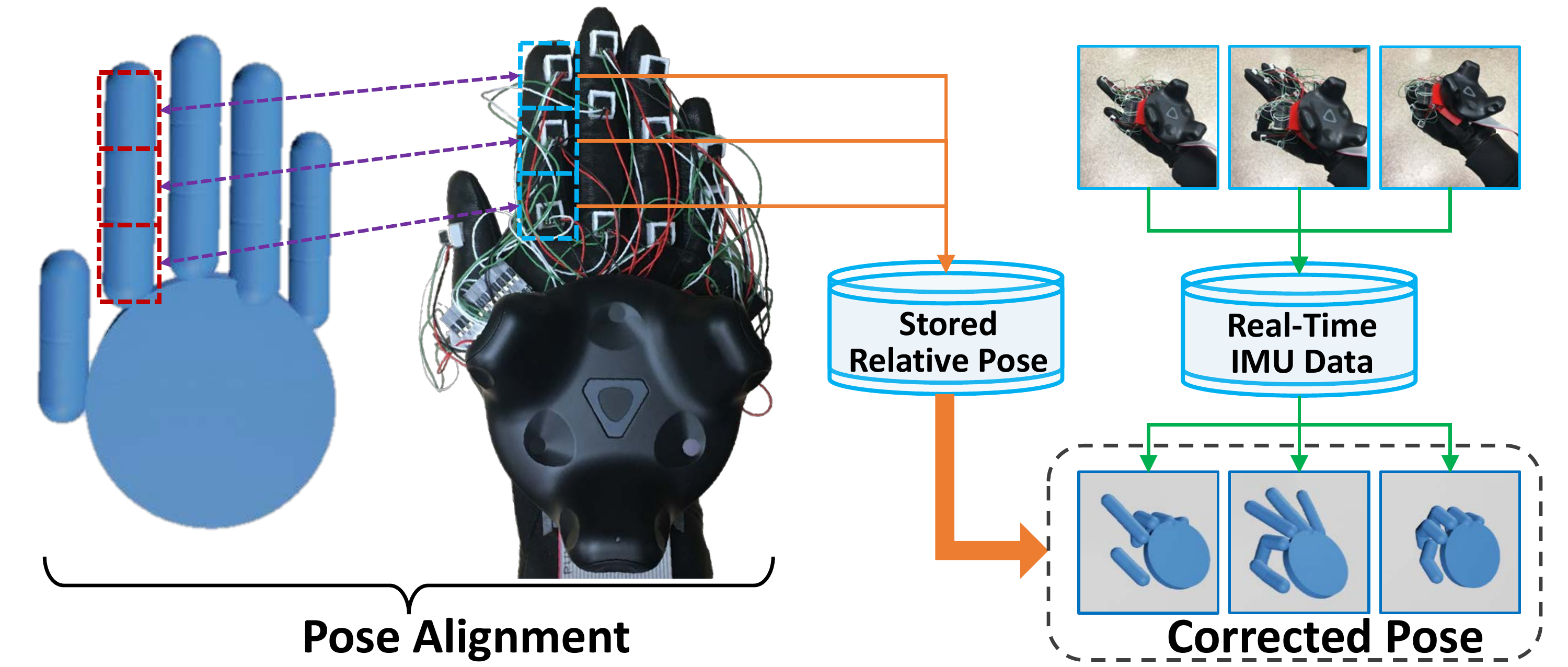}
    \caption{\textbf{The IMU calibration protocol.} The protocol starts by holding the hand flat, as shown by the virtual hand model. The relative pose between the world frame and the IMU's local coordinate system is recorded. The inverse of the recorded relative pose corrects the IMU data. Reproduced from Ref.~\cite{liu2019high} with permission.}
    \label{fig:gesture:calibration}
\end{figure}

Inertial sensors such as IMUs suffer from a common problem of drifting, which causes an accumulation of errors during operations. To overcome this issue, we introduce an IMU calibration protocol. When the sensed hand gesture degrades significantly, the user wearing the glove can hold the hand flat and maintain this gesture (see \cref{fig:gesture:calibration}) to initiate calibration; the system records the relative pose between the IMU and world frames. The orientation data measured by the IMUs are multiplied by the inverse of this relative pose to cancel out the differences, thus eliminating accumulated errors due to drifting. This routine can be performed conveniently when experiencing unreliable hand gesture sensing results.

\subsection{IMU evaluation}\label{sec:gesture:imu_eval}

We evaluated an individual IMU's bias and variance during rotations. Furthermore, we examined how accurately two articulated IMUs can reconstruct a static angle, indicating the performance of an atomic element in sensing the finger joint angle.

\paragraph{Evaluations of a single IMU}

\begin{figure}[t!]
    \centering
    \begin{subfigure}[b]{\linewidth}
        \centering
        \includegraphics[width=\linewidth,trim=1.05cm 0cm 1.35cm 0.3cm,clip]{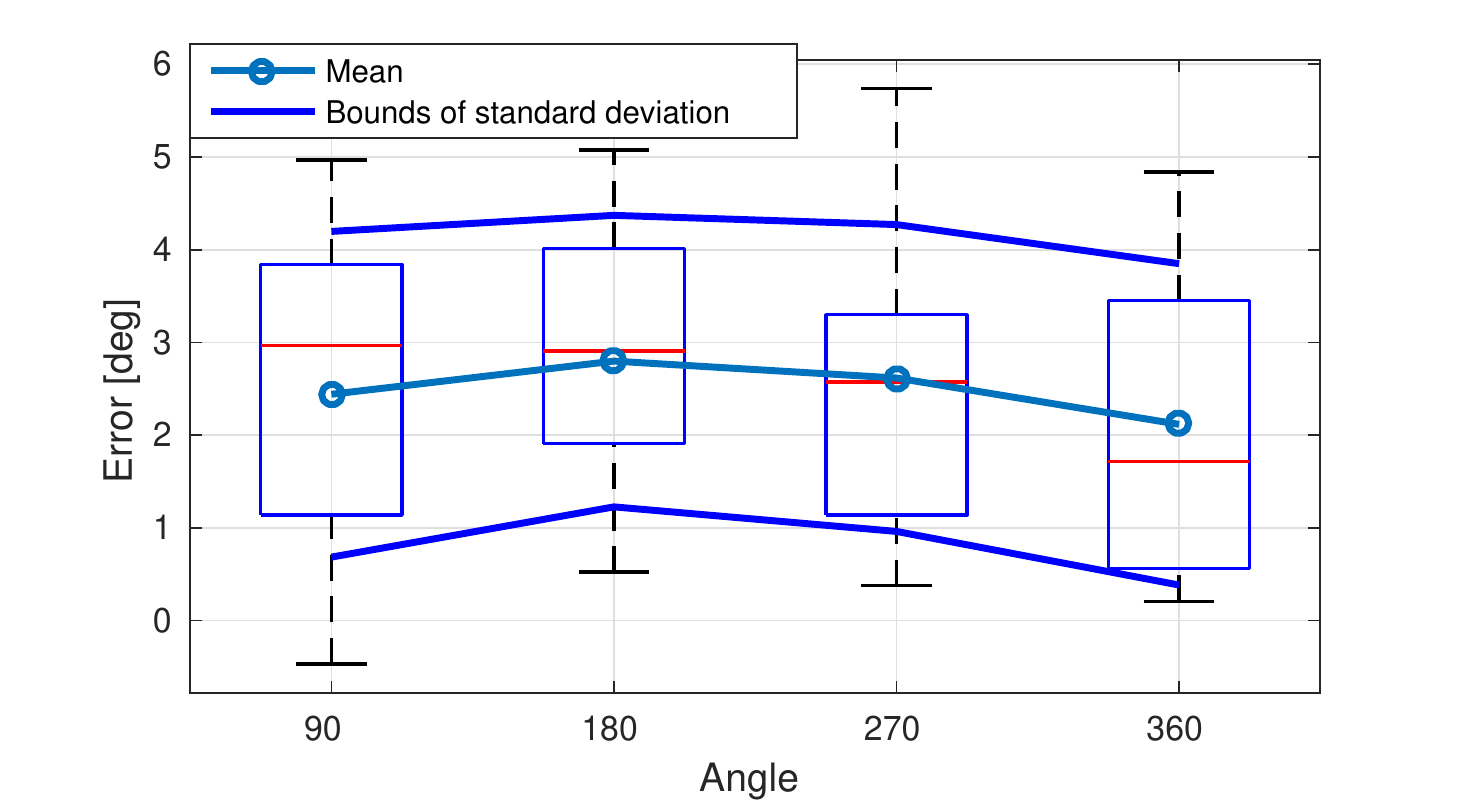}
        \caption{Error in measurement using single IMU.}
        \label{tab:gesture:imu_std}
    \end{subfigure}%
    \\
    \begin{subfigure}[b]{\linewidth}
        \centering
        \includegraphics[width=\linewidth,trim=0.86cm 0cm 1.35cm 0.3cm,clip]{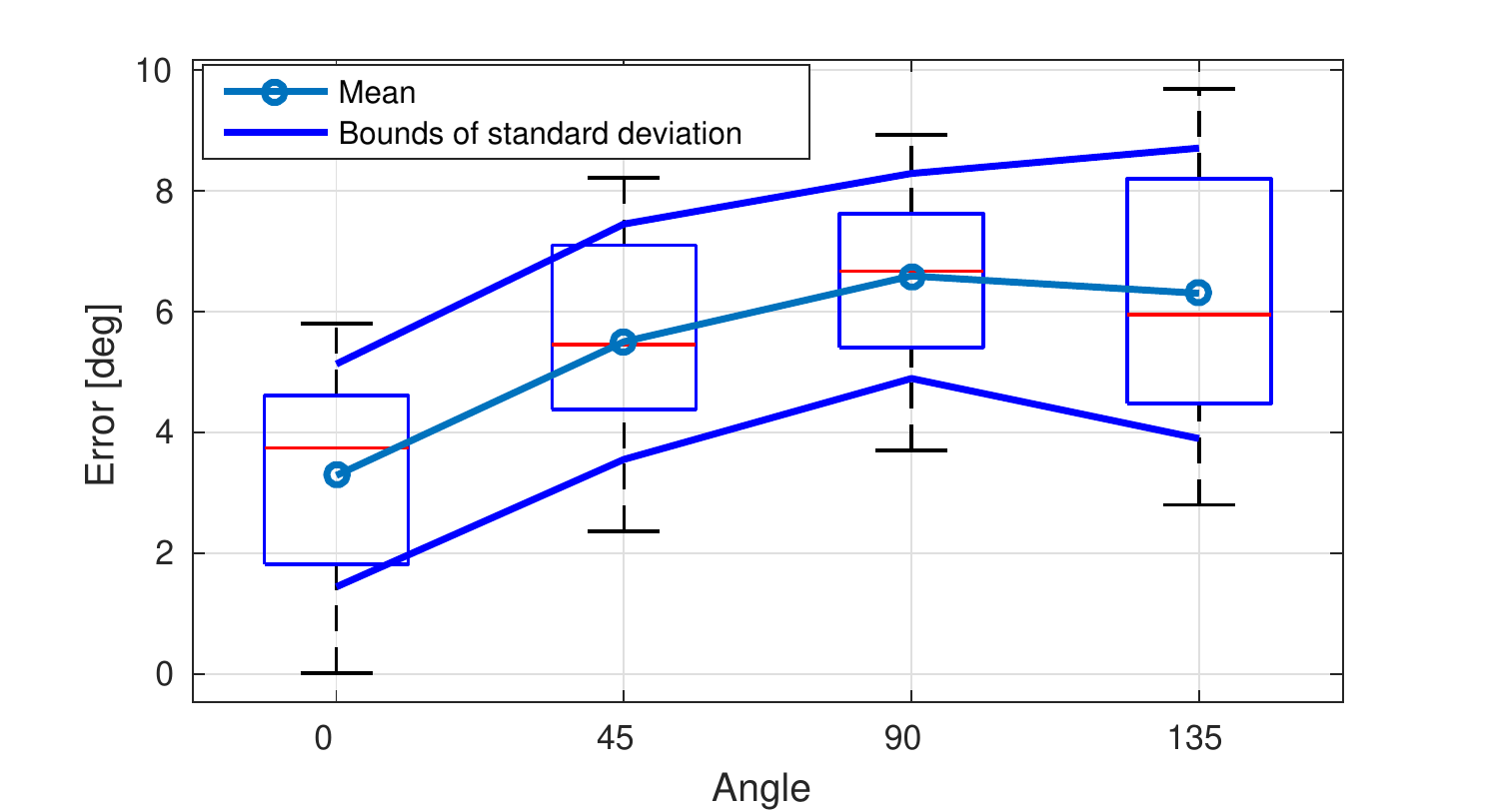}
        \caption{Error in recovering a fixed angle with two articulated IMUs.}
        \label{tab:gesture:imu_ex3}
    \end{subfigure}%
    \\
    \begin{subfigure}[b]{0.55\linewidth}
        \includegraphics[width=\linewidth]{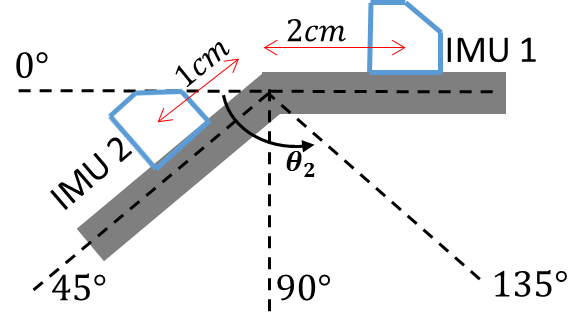}
        \caption{Schematic of articulated IMUs.}
        \label{tab:gesture:angle}
    \end{subfigure}%
    \hfill%
    \begin{subfigure}[b]{0.442\linewidth}
        \includegraphics[width=\linewidth]{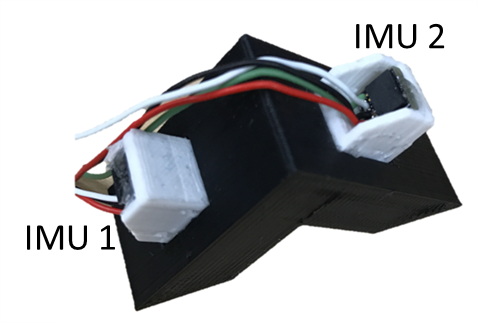}
        \caption{Physical setup.}
        \label{tab:gesture:angle_print}
    \end{subfigure}%
    \caption{\textbf{Evaluations of IMU performance.} The measurement error is summarized as the mean and standard deviation of (a) a single IMU and (b) two articulated IMUs under different settings. The red horizontal lines, blue boxes, and whiskers indicate the median error, the $25^{th}$ and $75^{th}$ percentiles, and the range of data points not considered to be outliers, respectively. A schematic of the experimental setup for evaluating the angle reconstruction with two articulated IMUs is shown in (c), and its physical setup with a $90^{\circ}$ bending angle is shown in (d). Reproduced from Ref.~\cite{liu2017glove} with permission.}
\end{figure}

As the reliability of the gesture sensing primarily depends on the IMU performance, it is crucial to investigate the IMU's bias and variance. More specifically, we rotated an IMU using a precise stepper motor controlled by an Arduino microcontroller. Four rotation angles--$90^{\circ}$, $180^{\circ}$, $270^{\circ}$, and $360^{\circ}$--were executed 20 times each at a constant angular velocity of 60 rotations per minute (RPMs). We did not test for a rotation angle exceeding $360^{\circ}$, as this is beyond the fingers' motion range. \cref{tab:gesture:imu_std} summarizes the mean and the standard deviation of the measured angular error. Overall, the IMU performed consistently with a bias between $2^{\circ}$ and $3^{\circ}$ and a $\pm 1.7^{\circ}$ standard deviation, suggesting that post-processing could effectively reduce the sensor bias.

\paragraph{Evaluations of articulated IMUs}

Evaluating IMU performance on whole-hand gesture sensing is difficult due to the lack of ground truth. As a compromise, we 3D printed four rigid bends with angles of $0^{\circ}$ $45^{\circ}$, $90^{\circ}$, and $135^{\circ}$ to emulate four specific states of finger bending, which evenly divided a finger joint's motion range as defined in \cref{eq:joint_limit_F}. Using two IMUs to construct a bend, assuming it to be a revolute joint, we tested the accuracy of the reconstructed joint angle by computing the relative poses between the two IMUs. \cref{tab:gesture:angle} shows a schematic of this experimental setup, and \cref{tab:gesture:angle_print} shows the physical setup with a $90^{\circ}$ bending angle. During the test, one IMU was placed $2~cm$ behind the bend, and another was placed $1~cm$ ahead, simulating the IMUs attached to a proximal phalanx and a middle phalanx, respectively. We repeated the test 20 times for each rigid bend. \cref{tab:gesture:imu_ex3} shows the errors of the estimated joint angles. As the bending angle increased, the reconstruction errors increased from $4^{\circ}$ to about $6^{\circ}$, with a slightly expanded confidence interval. Overall, the errors were still reasonable, although the IMUs tended to underperform as the bending angle increased. Through combination with the pose calibration protocol, these errors can be better counterbalanced, and the utilized IMU network can reliably support the collection of grasping data (see \cref{sec:application} for various case studies).

\section{Tactile-sensing mode}\label{sec:tactile}

Our reconfigurable data glove can be easily configured to the tactile-sensing mode, which shares the unified backbone design described in \cref{sec:gesture}. The tactile-sensing mode measures the distribution of forces exerted by the hand during complex hand-object interactions. We start by describing the force sensor specifications in \cref{sec:tactile_sensor}, which is followed by details of prototyping in \cref{sec:tactile_prototype}. We conclude this section with a qualitative evaluation in \cref{sec:tactile_eval}.

\begin{figure}[t!]
    \centering
    \begin{subfigure}[b]{0.482\linewidth}
        \includegraphics[width=\linewidth]{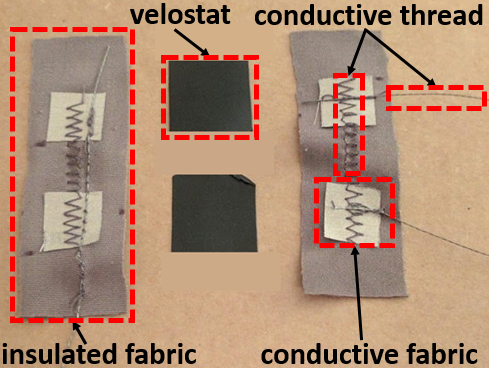}
        \caption{Velostat sensor construction}
        \label{tab:tactile:velostat}
    \end{subfigure}%
    \begin{subfigure}[b]{0.51\linewidth}
        \includegraphics[width=\linewidth]{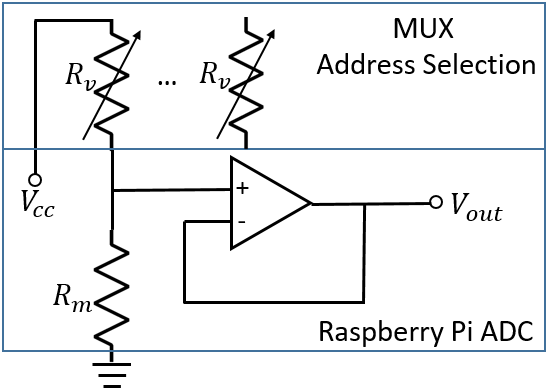}
        \caption{Velostat sensor circuit}
        \label{tab:tactile:circuit}
    \end{subfigure}%
    \\
    \begin{subfigure}[b]{\linewidth}
        \includegraphics[width=\linewidth,trim=0.5cm 0cm 1cm 0cm,clip]{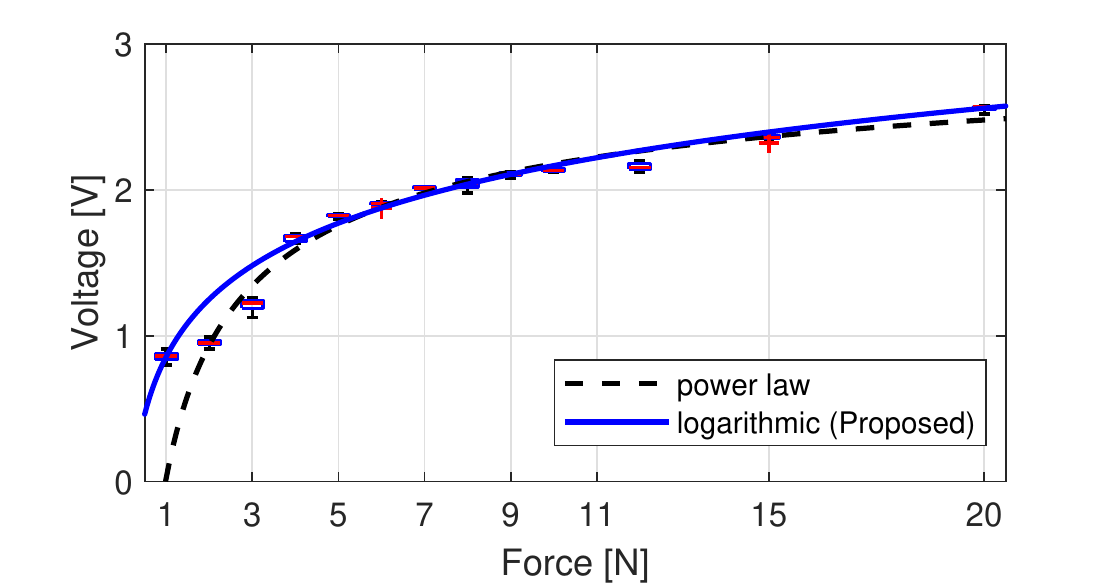}
        \caption{Force-voltage relation of one constructed Velostat sensing taxel}
        \label{tab:tactile:calib}
    \end{subfigure}%
    \\
    \begin{subfigure}[b]{0.196\linewidth}
        \centering
        \includegraphics[width=\linewidth]{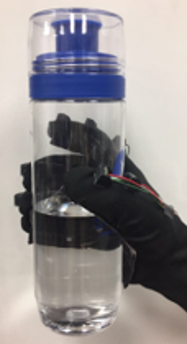}
        \caption{}
        \label{tab:tactile:grasping}
    \end{subfigure}%
    \begin{subfigure}[b]{0.8\linewidth}
        \centering
        \includegraphics[width=\linewidth]{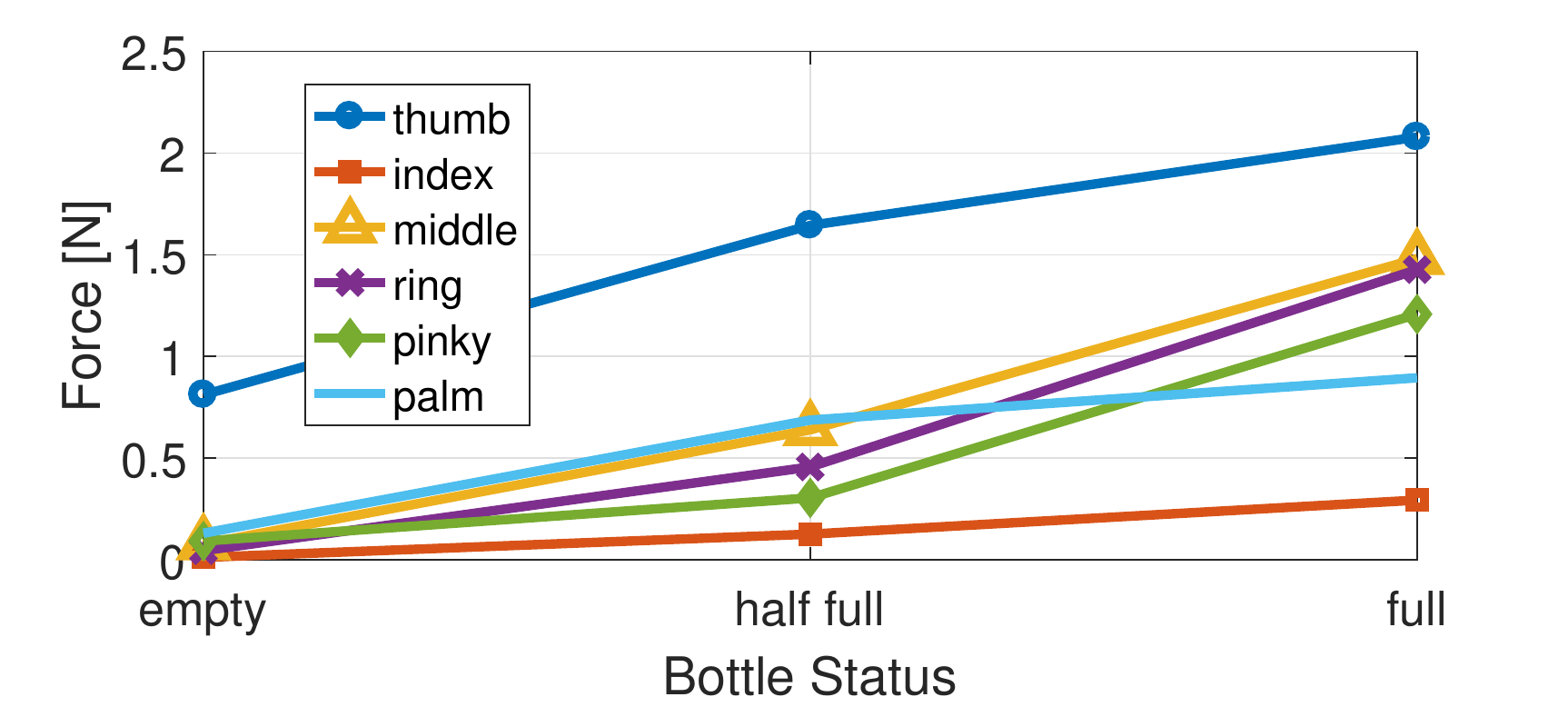}
        \caption{}
        \label{tab:tactile:force_velo}
    \end{subfigure}%
    \caption{\textbf{Characterization of the Velostat force sensor.} (a) The multi-layer structure of a Velostat force sensor. (b) The circuit layout for force data acquisition. (c) The force-voltage relation of one sensing taxel. Instead of using a power law, our choice of a logarithmic law fits the data better. (d) A grasp of the half-full bottle. (e) Force responses of grasping empty, half-full, and full bottles, respectively. Reproduced from Ref.~\cite{liu2017glove} with permission.}
\end{figure}

\subsection{Force sensor}\label{sec:tactile_sensor}

We adopt a network of force sensors made from Velostat to provide force sensing in this tactile-sensing mode. \cref{tab:tactile:velostat} illustrates the Velostat force sensor's multi-layer structure. A taxel (\ie, a single-point force-sensing unit) is composed of one inner layer of Velostat ($2~cm \times 2~cm$) and two middle layers of conductive fabric, stitched together by conductive thread and enclosed by two outer layers of insulated fabric. A force-sensor pad consisting of 2 taxels is placed on each finger, and a sensor grid with $4~cm \times 4~cm$ taxels is placed on the palm. Lead wires to the pads and grid are braided into the conductive thread.

As the Velostat's resistance changes with different pressing forces, the measured voltage across a taxel can be regarded as the force reading at that region. To acquire the voltage readings, we connect these Velostat force-sensing taxels in parallel via analog multiplexers controlled by the RPi's GPIO and output to its SPI-enabled ADS1256 ADC. More specifically, two 74HC4051 multiplexers are used for the palm grid, and a CD74HC4067 multiplexer is used for all the finger pads. A voltage divider circuit, shown in \cref{tab:tactile:circuit}, is constructed by connecting a $200\Omega$ resistor between the RPi's ADC input channel and the multiplexers.

We now characterize the sensor's force-voltage relation~\cite{lee2016feasibility}. A total of 13 standard weights ($0.1~kg$ to $1.0~kg$ with $0.1~kg$ increments, $1.2~kg$, $1.5~kg$, and $2.0~kg$) were applied to a taxel, and the associated voltages across that taxel were measured. The calibration circuit was the same as that in \cref{tab:tactile:circuit}, except that only the taxel of interest was connected. The weights in kilograms were converted to forces in Newtons with a gravitational acceleration $g=10~m/s^2$. We first tested the power law~\cite{lee2016feasibility} for characterizing the force-voltage relation of a taxel. The result was $F = -1.067V^{-0.4798}+3.244$ with $R^2 = 0.9704$, where $F$ is the applied force, and $V$ is the output voltage. However, we further tested a logarithmic law, resulting in a better force-voltage relation: $F = 0.569\log{(44.98V)}$ with a higher $R^2 = 0.9902$. Hence, we adopted the logarithmic fit to establish a correspondence between the voltage reading across a taxel and the force the taxel is subjected to. \cref{tab:tactile:calib} compares these two fits.

\subsection{Prototyping}\label{sec:tactile_prototype}

\cref{fig:summary}a~\cite{liu2017glove,liu2019high,li2020incremental} displays a prototype of the tactile-sensing glove. The capability of force sensing is accomplished by placing one Velostat force-sensing pad on each finger (one taxel in the proximal area and another in the distal area) and a single $4~cm \times 4~cm$ Velostat force-sensing grid over the glove's palm region. Based on the established force-voltage relation, these taxels collectively measure the distribution of forces exerted by the hand. Meanwhile, the 15 IMUs capture the hand gestures in motion. These components are all connected to the RPi, which can be remotely accessed to visualize and subsequently utilize the collected gesture and force data in a local workstation, providing a neat solution to collect human manipulation data.

By measuring the voltage and current across each component, we investigated the power consumption of the prototype. \cref{tab:tactile:power} reports the peak power of each component of interest as the product of its voltage and current in a 10-min operation. The total power consumption was $2.72~W$, which can be easily powered by a conventional Li-Po battery, offering an untethered user experience and natural interactions during data collection.

\begin{table}[htb!]
    \centering
    \small
    \setlength{\tabcolsep}{4pt}
	\caption{\textbf{Power consumption of the tactile-sensing glove.}}
	\label{tab:tactile:power}
    \begin{tabular}{ccccc}
        \toprule
        \multirow{2}{*}{\textbf{Component}} & gesture sensing & force sensing & computing & \multirow{2}{*}{total} \\
         & $15$ IMUs & $6$ Velostat & RPI & \\
        \midrule
        \textbf{Power (W)} & 0.60 & 0.02 & 2.15 & 2.72 \\
        \bottomrule
    \end{tabular}
\end{table}

\subsection{Qualitative evaluation}\label{sec:tactile_eval}

We evaluated the performance of the tactile-sensing glove in differentiating among low, medium, and high forces by grasping a water bottle in three states, \textit{empty}, \textit{half-full}, and \textit{full}, whose weights were $0.13~kg$, $0.46~kg$, and $0.75~kg$, respectively. The participants were asked to perform the grasps naturally and succinctly--exerting a force just enough to prevent the bottle from slipping out of the hand; \cref{tab:tactile:grasping} shows such an instance. Ten grasps were performed for each bottle state. To simplify the analysis, the force in the palm was the average of all 16 force readings of the palm grid, and the force in each finger was the average reading of the corresponding finger pads. \cref{tab:tactile:force_velo} shows the recorded forces exerted by different hand regions.

\section{VR mode}\label{sec:vr}

Since the different modes of our data glove share a unified backbone design, reconfiguring the glove to the VR mode in order to obtain contact points during interactions can be achieved with only three steps. First, given the sensed hand gestures obtained by the shared backbone, we need to construct a virtual hand model for interactions (see \cref{sec:vr_hand_model}). Next, we must develop an approach to achieve a stable grasp of virtual objects (see \cref{sec:vr_grasp}). Finally, grasping objects in VR introduces new difficulty without a tangible object being physically manipulated; we leverage haptic feedback to address this problem in \cref{sec:vr_haptic}. We conclude this section with an evaluation in \cref{sec:vr_eval}.

\subsection{Virtual hand model}\label{sec:vr_hand_model}

\begin{figure}[t!]
    \centering
    \begin{subfigure}[b]{0.333\linewidth}
        \includegraphics[width=\linewidth]{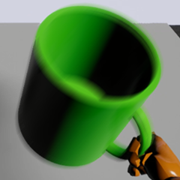}
        \caption{LeapMotion}
        \label{fig:vr:comparison_vision}
    \end{subfigure}%
    \begin{subfigure}[b]{0.333\linewidth}
        \includegraphics[width=\linewidth]{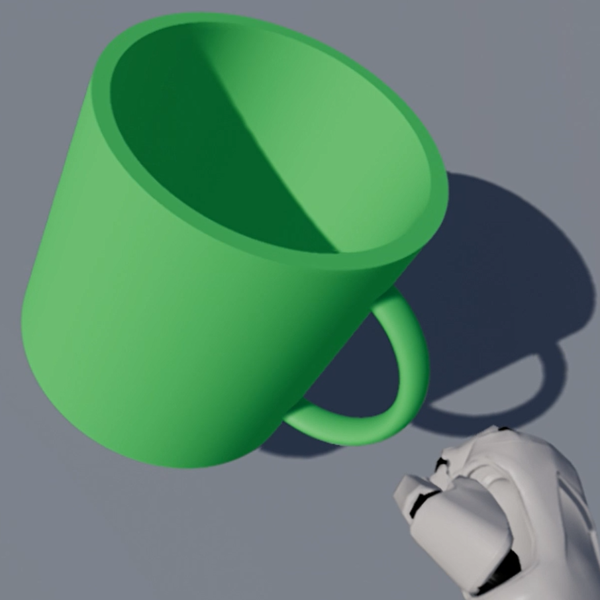}
        \caption{Oculus Touch}
        \label{fig:vr:comparison_touch}
    \end{subfigure}%
    \begin{subfigure}[b]{0.333\linewidth}
        \includegraphics[width=\linewidth]{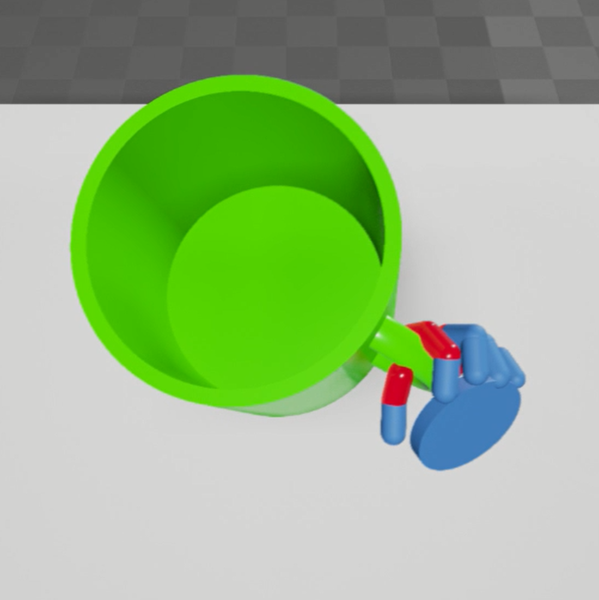}
        \caption{Ours}
        \label{fig:vr:comparison_glove}
    \end{subfigure}%
    \caption{\textbf{A comparison of grasp among (a) a LeapMotion sensor, (b) an Oculus Touch controller, and (c) our reconfigurable glove system in the VR mode.} The grasp in (a) is unstable, as reflected by the motion blur, due to occlusion in the vision-based hand gesture sensing approach. While (b) affords a form of ``stable'' grasp (\ie, it removes the gravity from the cup) by directly attaching the object to the hand, this approach is unnatural, with minimal realism. It does not reflect the actual contact between a hand and an object, and sometimes the hand even fails to come into contact with the object. (c) The proposed reconfigurable glove in VR mode offers a realistic and stable grasp, which is crucial for obtaining contact points during interactions. Reproduced from Ref.~\cite{liu2019high} with permission.}
    \label{fig:vr:comparison}
\end{figure}

Generating a stable grasp is the prerequisite for obtaining contact points during interactions. Existing vision-based hand gesture sensing solutions, including commercial projects such as LeapMotion~\cite{leapmotion} and RealSense~\cite{realsense}, struggle with stable grasps due to occlusions, sensor noises, and a limited field of view (FoV); interested readers can refer to \cref{fig:vr:comparison_vision} for a comparison in a typical scenario. In comparison, existing VR controllers adopt an alternative approach--the virtual objects are directly attached to the virtual hand when a grasp event is triggered. As illustrated in \cref{fig:vr:comparison_touch}, the resulting experience has minimal realism and cannot reflect the actual contact configuration. The above limitations motivate us to realize a stable virtual grasp by developing a caging-based approach that is capable of real-time computation while offering sufficient realism; an example is provided in \cref{fig:vr:comparison_glove}.

\begin{figure}[t!]
    \centering
    \includegraphics[width=0.8\linewidth]{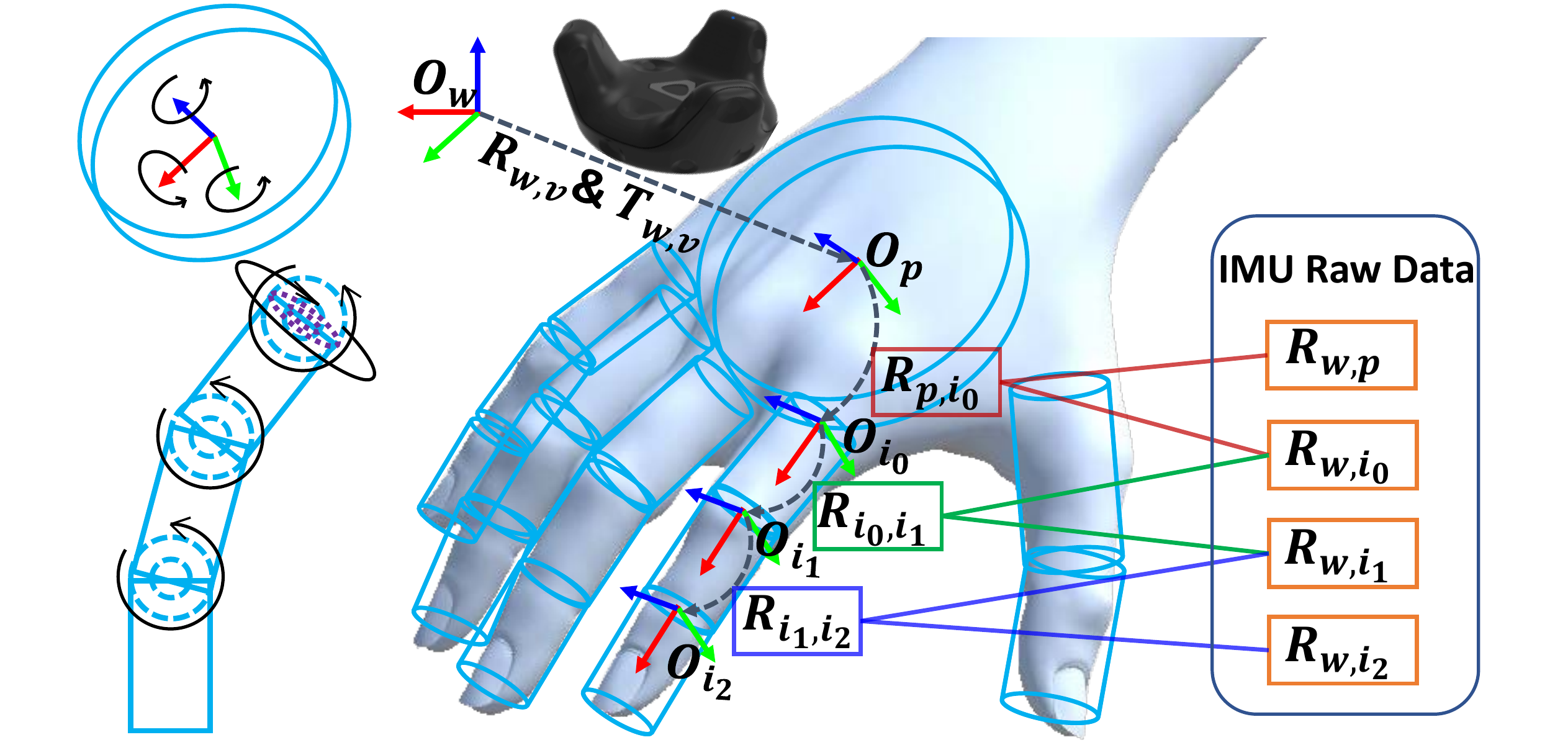}
    \caption{\textbf{Structure of the virtual hand model}. Each phalanx is modeled by a small cylinder whose dimensions are measured by a participant. The pose of each phalanx is reconstructed from the data read by the IMUs. The Vive tracker provides direct tracking of the hand pose. Reproduced from Ref.~\cite{liu2019high} with permission.}
    \label{fig:vr:hand}
\end{figure}

Thanks to the reconfigurable nature of the glove, creating a virtual hand model in VR is simply the reiteration of the hand gesture-sensing module described in \cref{sec:gesture}; \cref{fig:vr:hand} shows the structure of the virtual hand. More specifically, the hand gestures in the local frames are given by the IMUs, and a Vive tracker with HTC Lighthouse provides the precise positioning of the hand in a global coordinate, computed by the time-difference-of-arrival.

\subsection{Stable grasps}\label{sec:vr_grasp}

Methods for realizing a virtual grasp in VR can be roughly categorized into two streams, with their unique pros and cons. One approach is to use a physics-based simulation with collision detection to support realistic manipulations by simulating the contact between a soft hand and a virtual object made from varied materials. Despite its high fidelity, this approach often demands a significant amount of computation, making it difficult--if not impossible--to use in real time. Alternatively, symbolic-based and rule-based grasps are popular approaches. A grasp or release is triggered based on a set of predefined rules when specific conditions are satisfied. This approach is computationally efficient but provides minimal realism.

Our configurable glove-based system must balance the above two factors to obtain contact points during interactions. It must provide a more natural interaction than those of rule-based methods, such that the contact points obtained on the objects are relatively accurate, while ensuring more effective computation than high-fidelity physics-based simulations, such that it can be achieved in real time.

\begin{figure}[t!]
    \centering
    \includegraphics[width=\linewidth]{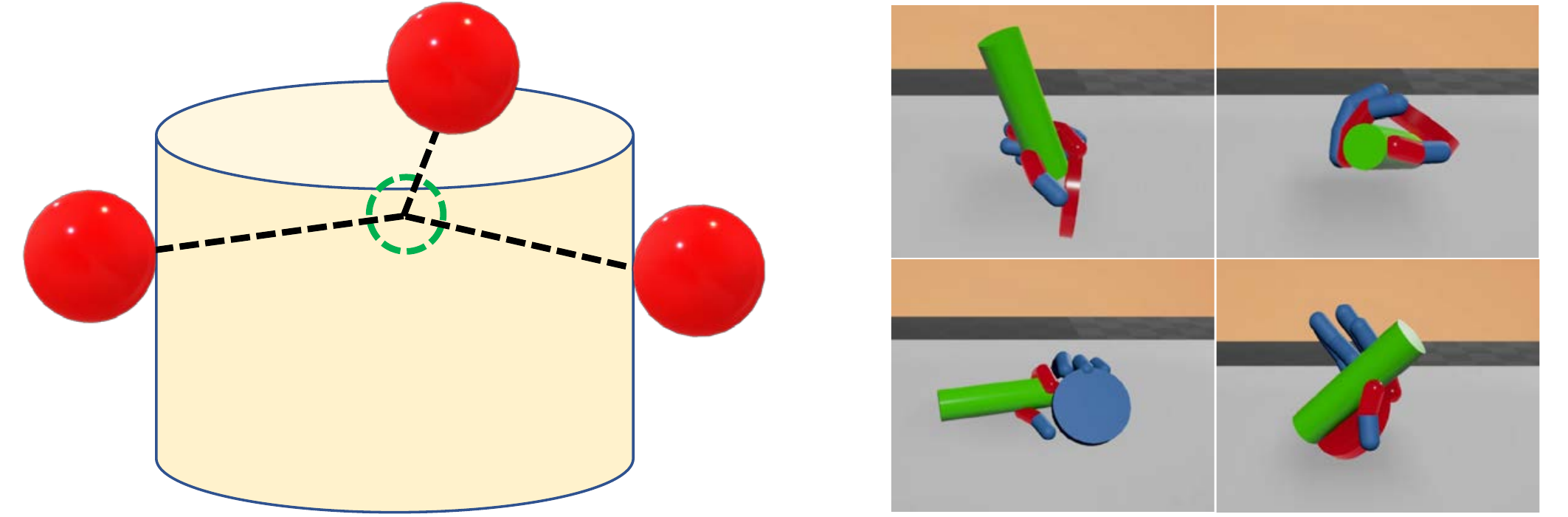}
    \begin{tabular}{c}
        \hspace{0.2cm}(a) \hspace{3.8cm}(b)
    \end{tabular}
    \caption{\textbf{Detect stable grasps based on collisions.} (a) When the geometric center (green dashed circle) of all the collision points (red balls) overlaps with the object (yellow cylinder), the object is considered to be stably grasped and will move along with the hand. (b) Various stable grasps of a small green cylinder. Reproduced from Ref.~\cite{liu2019high} with permission.}
    \label{fig:vr:collision}
\end{figure}

In this work, we devise a caging-based stable grasp algorithm, which can be summarized as follows. First, the algorithm detects all collisions between the hands and objects \eg, the red areas in \cref{fig:vr:collision}b). Next, the algorithm computes the geometric center of all collision points between the hands and objects and checks whether this center is within the object. Supposing that the above situation holds (see \cref{fig:vr:collision}a), we consider this object to be ``caged''; thus, it can be stably grasped. The objects' physical properties are turned off, allowing them to move along with the hand. Otherwise, only standard collisions are triggered between the hand and object. Finally, the grasped object is released when the collision event ends or the geometric center of the collisions is outside the object. This process ensures that a grasp only starts after a caging is formed, offering a more natural manipulation experience with higher realism than rule-based grasps.

\subsection{Haptic feedback}\label{sec:vr_haptic}

By default, the participants have no way to feel whether or not their \textit{virtual} hands are in contact with the \textit{virtual} objects while operating the glove in VR mode due to the lack of haptic feedback, which prevents they from manipulating objects naturally. To fill this gap, the VR mode implements a network of shaftless vibration motors that are triggered when the corresponding virtual phalanxes collide with the virtual object; this offers an effective means of providing each finger with vibrational haptic feedback in the \textbf{physical} world that corresponds to the contact feedback that the participants should receive in VR. Connected to a 74HC4051 analog multiplexer and controlled by the RPi's GPIO, these small ($10~mm \times 2~mm$) and lightweight ($0.8~g$) vibration motors provide $14,500$ RPM with a $3~V$ input voltage. Once a finger touches the virtual object, the vibration motors located at that region of the glove are activated to provide continuous feedback. When the hand forms a stable grasp, all motors are powered up, so that the user can maintain the current hand gesture to hold the object.

\subsection{Qualitative evaluation}\label{sec:vr_eval}

\begin{figure}[t!]
    \centering
    \begin{overpic}
        [width=\linewidth]{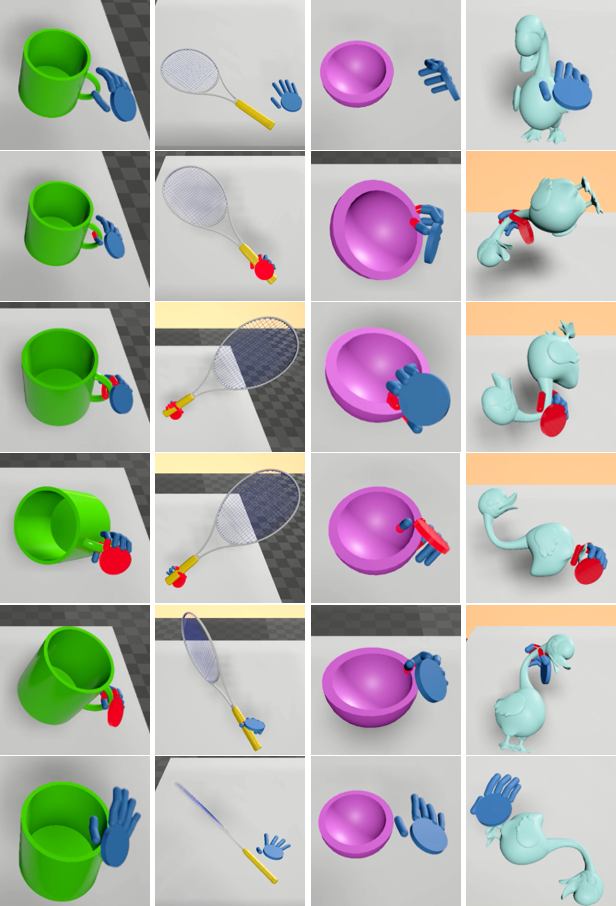}
        \put(8,-3){\color{black} (a)}
        \put(24,-3){\color{black} (b)}
        \put(42,-3){\color{black} (c)}
        \put(58,-3){\color{black} (d)}
    \end{overpic}
    \vspace{12pt}
    \caption{\textbf{Various grasp results for four virtual objects:} (a) a mug, (b) a tennis racket, (c) a bowl, and (d) a goose toy. The top and bottom rows show the approach and release of the target objects, respectively. Reproduced from Ref.~\cite{liu2019high} with permission.}
    \label{fig:vr:grasp_result}
\end{figure}
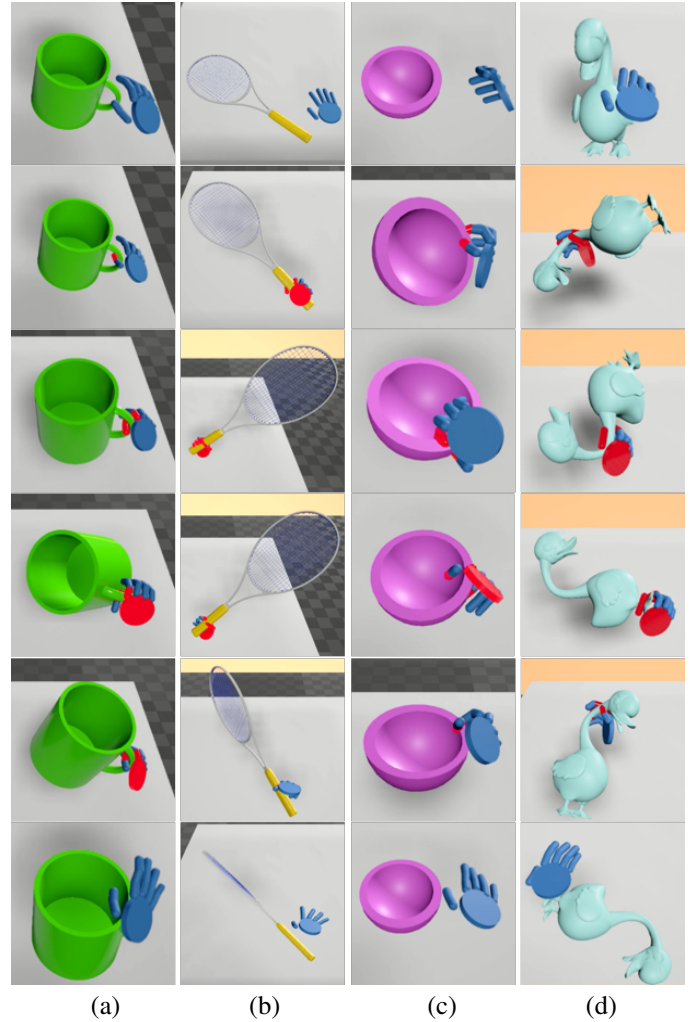

We conducted a case study wherein the participants were asked to wear the VR glove and grasp four virtual objects with different shapes and functions, including a mug, a tennis racket, a bowl, and a goose toy (see \cref{fig:vr:grasp_result}). These four objects were selected because (i) they are everyday objects with a large variation in their geometry, providing a more comprehensive assessment of the virtual grasp; and (ii) each of the four objects can be grasped in different manners based on their functions, covering more grasp types~\cite{feix2016grasp,liu2021synthesizing}. We started by testing different ways of interacting with virtual objects, such as grasping a mug by either the handle or the rim. Such diverse interactions afforded a natural experience by integrating unconstrained fine-grained gestures, which is difficult for existing platforms (\eg, LeapMotion). In comparison, our reconfigurable glove in VR mode successfully balanced the naturalness of the interactions with the stability of the grasp, providing a better realism in VR, which was close to how objects are manipulated in the physical world.

Notably, the reconfigurable glove in VR mode was able to track hand gestures and maintain a stable grasp even when the hand was outside the participant's FoV, thus offering a significant advantage compared with vision-based approaches (\eg, the LeapMotion sensor). In a comparative study in which the participant's hand could be outside of the FoV, the performance using the VR glove significantly surpassed that of LeapMotion (see \cref{tab:vr:result}), thereby demonstrating the efficacy of the VR glove hardware, the caging-based grasp approach, and the haptic feedback.

\begin{table}[htb!]
    \centering
    \small
    \caption{\textbf{Success rates of grasping and moving four different objects using the VR glove (G) and the LeapMotion sensor (L)}.}
    \label{tab:vr:result}
    \begin{tabular}{c c c c c c}
        \toprule
        \textbf{task} & \textbf{setup} & \textbf{mug} & \textbf{racket} & \textbf{mug} & \textbf{racket}\\
        \midrule
        \multirow{2}{*}{grasp} & L & 80\% & 13\% & 27\% & 67\% \\
        & G & 100\% & 100\% & 100\% & 93\% \\
        \multirow{2}{*}{move} & L & 33\% & 7\% & 0\% & 47\% \\
        & G & 100\% & 93\% & 93\% & 87\% \\
        \bottomrule
    \end{tabular}
\end{table}

\section{Simulation mode}\label{sec:sim}

A manipulation event consists of both hand information and object information. Most prior work has focused on the former without paying much attention to the latter. In fact, objects may be occluded or may even change significantly in shape as a result of a manipulation event, such as through deformation or cracking. Such information is essential in understanding the manipulation event, as it reflects the goals. However, existing solutions, even those with specialized sensors, fall short in handling this scenario, so a solution beyond the conventional scope of data gloves is called for.

To tackle this challenge, we integrate a state-of-the-art FEM simulator~\cite{li2020incremental} to reconstruct the physical effects of an object, in numeric terms, during the manipulation. Given the trajectory data obtained by the proposed glove-based system, both physical and virtual properties and how they evolve over time are simulated and rendered, providing a new dimension for understanding complex manipulation events.

\subsection{Simulation method}

We start with a brief background of solid simulation. Solid simulation is often conducted with FEM~\cite{zienkiewicz2000finite}, which discretizes each object into small elements with a discrete set of sample points as the DoFs. Then, mass and momentum conservation equations are discretized on the mesh and integrated over time to capture the dynamics, in which \textit{elasticity} and \textit{contact} are the most essential yet most challenging components. \textit{Elasticity} is the ability of an object to retain its rest shape under external impulses or forces, whereas \textit{contact} describes the intersection-free constraints on an object's motion trajectory. However, elasticity is nonlinear and non-convex, and contact is non-smooth, both of which can pose significant difficulties to traditional solid simulators based on numerical methods~\cite{li2020robust}. Recently, Li \etal~\cite{li2020incremental} proposed incremental potential contact (IPC), a robust and accurate contact-handling method for FEM simulations~\cite{li2020codimensional,fang2021guaranteed,ferguson2021intersection,lan2021medial,choo2021barrier}; it formulates the non-smooth contact condition into smooth approximate barrier potentials so that the non-smooth contact condition can be solved simultaneously with electrodynamics using a line search method~\cite{li2019decomposed,wang2020hierarchical,nocedal2006numerical} with a global convergence guarantee. As it is able to consistently produce high-quality results without numerical instability issues, IPC makes it possible to conveniently simulate complex manipulation events, even with extremely large deformations.

We further extend the original IPC to support object fracture by measuring the displacement
of \textit{every} pair of points; that is, we go through all pairs of points for a triangle and all triangles on the mesh. If the displacement relative to the pair of points' original distance exceeds a certain strain threshold (in this work, we set it to $1.1$), we mark the triangle in between as separated. At the end of every time step, we reconstruct the mesh topology using a graph-based approach~\cite{hegemann2013level}, according to the tetrahedra face separation information. Due to the existence of the IPC barrier, which only allows a positive distance between surface primitives, it is essential to ensure that, after the topology change, the split faces do not exactly overlap. Therefore, we perturb the duplicate nodes on the split faces by a tiny displacement toward the normal direction, which works nicely even when edge-edge contact pairs are ignored for simplicity.

\begin{figure*}[t!]
    \centering
    \begin{subfigure}[b]{0.25\linewidth}
        \includegraphics[width=\linewidth]{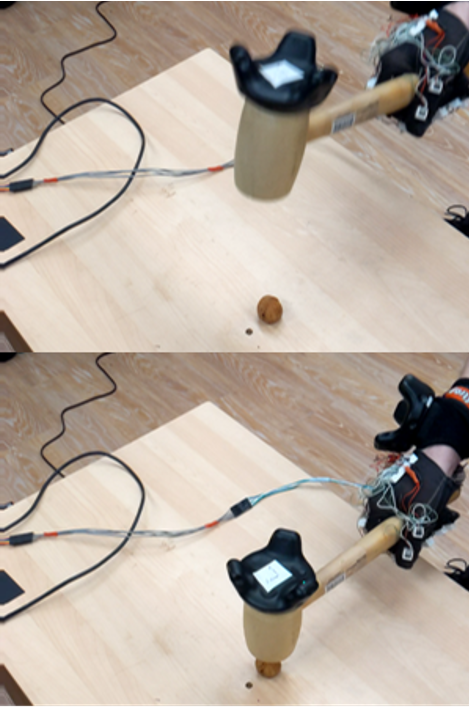}
        \caption{uncracked}
    \end{subfigure}%
    \begin{subfigure}[b]{0.25\linewidth}
        \includegraphics[width=\linewidth]{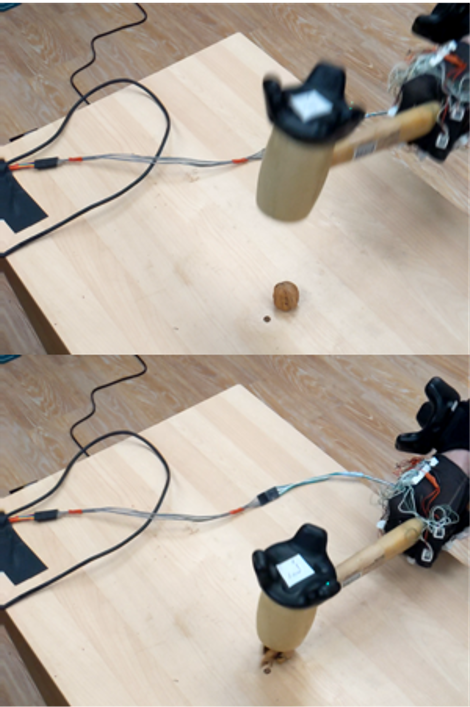}
        \caption{cracked}
    \end{subfigure}%
    \begin{subfigure}[b]{0.25\linewidth}
        \includegraphics[width=\linewidth]{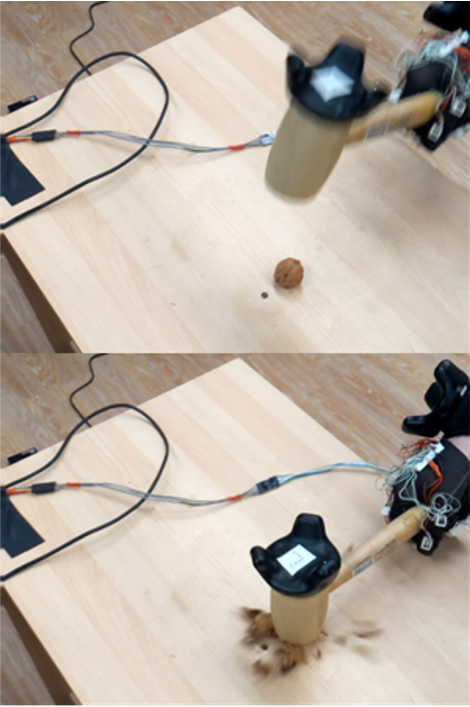}
        \caption{smashed}
    \end{subfigure}%
    \begin{subfigure}[b]{0.25\linewidth}
        \includegraphics[width=\linewidth]{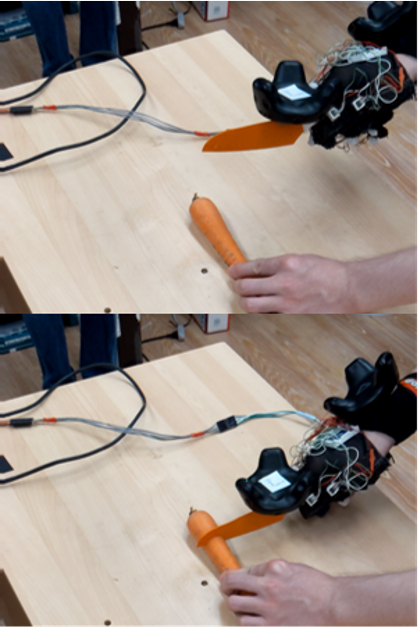}
        \caption{cut in half}
    \end{subfigure}%
    \caption{\textbf{Four types of tool-use events captured by a slow motion camera at 120 FPS.}}
    \label{fig:sim:slow_motion}
\end{figure*}

\begin{figure*}[b!]
    \centering
    \begin{subfigure}[b]{0.33\linewidth}
        \includegraphics[width=\linewidth]{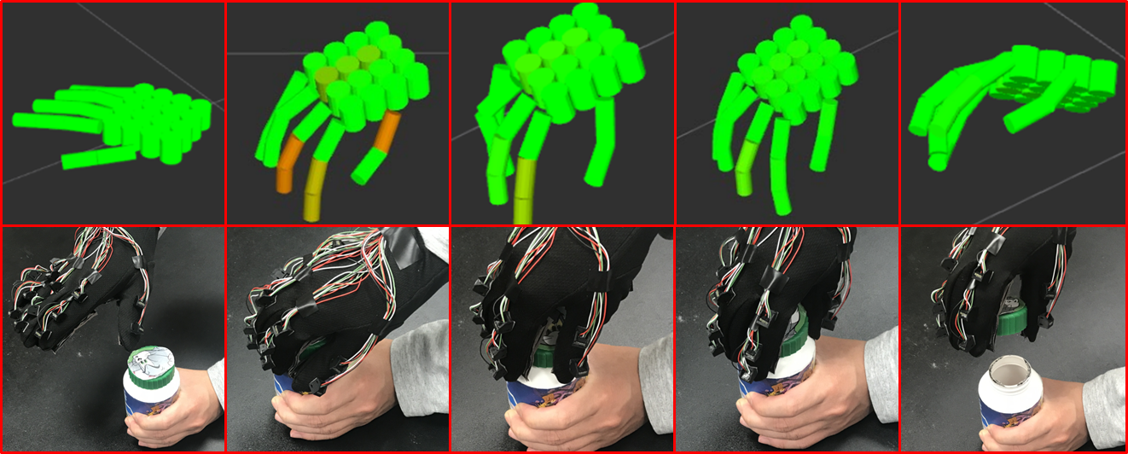}
        \caption{\textit{Bottle~1}, no childproof lock}
        \label{fig:69}
    \end{subfigure}%
    \hfill%
    \begin{subfigure}[b]{0.33\linewidth}
        \includegraphics[width=\linewidth]{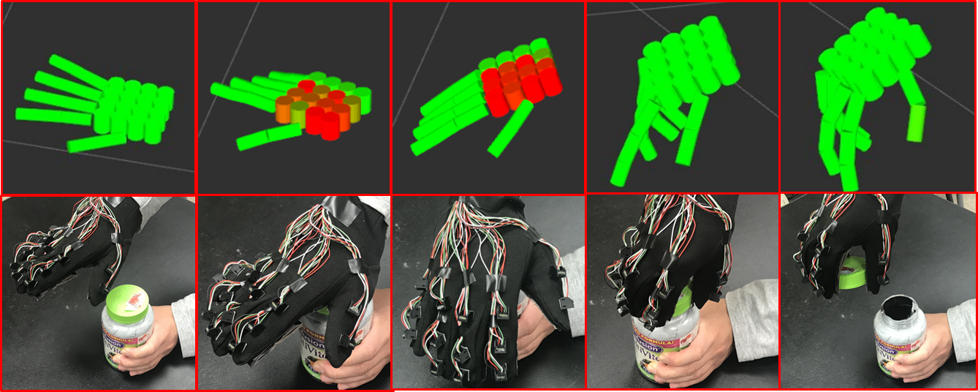}
        \caption{\textit{Bottle~2}, pressing down the lid to unlock}
        \label{fig:64}
    \end{subfigure}%
    \hfill%
    \begin{subfigure}[b]{0.33\linewidth}
        \includegraphics[width=\linewidth]{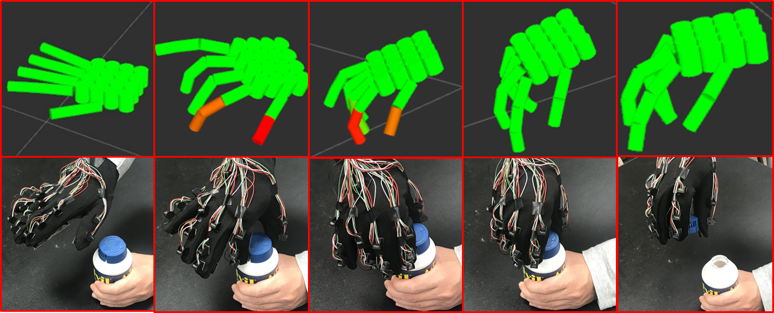}
        \caption{\textit{Bottle~3}, pinching the lid to unlock}
        \label{fig:68}
    \end{subfigure}%
    \caption{\textbf{Visualizations of the hand gesture and force of opening three bottles collected using the tactile-sensing glove.} These visualizations reveal the subtle differences between the actions of opening medicine bottles and opening a conventional bottle; the essence of this task is that visual information alone is insufficient to distinguish between the opening of the various bottles. Reproduced from Ref.~\cite{liu2017glove} with permission.}
    \label{fig:app:tactile:vis}
\end{figure*}

\subsection{Prototyping and input data collection}

The simulation-augmented glove-based system is essentially the same as the VR glove, except for the lack of vibration motors; however, it is augmented with the simulated force evolved over time. Compared with the aforementioned two hardware-focused designs, the simulation-augmented glove-based system offers an in-depth prediction of physics with fine-grained object dynamics—that is, how the geometry (\eg, large deformation) and topology (\eg, fracture) evolve. To showcase the efficacy of this system, we focus on a tool-use setting wherein a user manipulates a tool (\eg, a hammer) to apply on a target object (\eg, a nut), causing geometry and/or topology changes. To collect one set of data, the hand gestures and poses are reconstructed similarly using the other two glove-base systems. The tool's movement is further tracked to simulate the interactions between the tool and the object.

More specifically, two Vive trackers track the movements of the glove-based system (\ie, the hand) and the tool, respectively. The third tracker, which serves as the reference point for the target object (\eg, a nut) is fixed to the table. All three Vive trackers are calibrated such that their relative poses and the captured trajectories can be expressed in the same coordinate. The target objects and the tool's meshes are scanned beforehand using a depth camera. By combining the scanned meshes and captured trajectories, we can fully reconstruct a sequence of 3D meshes representing the movements of the hand and tool and simulate the resulting physical effects of the target object. The captured mesh sequences are directly input to the simulation as boundary conditions, and the DoFs being simulated are primarily those on the target object. \cref{fig:sim:slow_motion} shows some keyframes of the data collection for cracking walnuts and cutting carrots. It should be noted that capturing how the object changes and its physical properties over time is extremely challenging--if not impossible--using visual information alone.

\subsection{Simulation setup}

An object's material properties in a simulation are mainly reflected by its stiffness (\ie, the object is more difficult to deform or fracture if it is stiffer), governed by its Young's modulus and Poisson's ratio. These parameters must be set appropriately in the simulation in order to produce effects that match those in the physical world. The Young's modulus and Poisson's ratio of a material can be found in related works~\cite{bourne2002food,williams2005mechanical,kiani2011determination}. Another parameter that must be set is the fracturing strain threshold, which determines the dimension of the segments when fracturing is triggered. This parameter is tuned so that the simulator can reproduce the type of effects observed in the physical world. The time step of the simulation is the inversion of the sampling frequency of the Vive trackers that acquire the trajectories.

\section{Applications}\label{sec:application}

In this section, we showcase a series of applications by reconfiguring the data glove to the tactile-sensing mode (\cref{sec:app:tactile}), VR mode (\cref{sec:app:vr}), and simulation mode (\cref{sec:app:sim}), all of which share the same backbone design. (Interested readers can also refer to the Appendix for video demonstrations.)

\subsection{Tactile-sensing mode}\label{sec:app:tactile}

We evaluated the tactile-sensing mode by capturing the manipulation data of opening three types of medicine bottles. Two of these bottles are equipped with different locking mechanisms and require a series of specific action sequences to remove the lid. More specifically, \textit{Bottle 1} does not have a safety lock, and simply twisting the lid is sufficient to open it. The lid of \textit{Bottle 2} must be pressed simultaneously while twisting it. \textit{Bottle 3} has a safety lock in its lid, which requires a pinching action before twisting to unlock it. Notably, the pressing and pinching actions required to open \textit{Bottle 2} and \textit{Bottle 3} are challenging to recognize without using the force information recorded by the glove.

\cref{fig:app:tactile:vis} shows examples of the recorded data with both hand gesture and force information. The first row of \cref{fig:app:tactile:vis} visualizes the captured manipulation action sequences of opening these three bottles. The second row shows the corresponding action sequences captured by an RGB camera for reference.

Qualitatively, compared with the action sequences shown in the second row, the visualization results in the first row differentiate the fine manipulation actions with additional force information. For example, the fingers in \cref{fig:64} are flat and parallel to the bottle lid, whereas those in \cref{fig:68} are similar to those in the gripping pose. The responses of the force markers are also different due to varying contact points between the human hand and the lid: The high responses in \cref{fig:64} are concentrated on the palm area, whereas only two evident responses on the distal thumb and index finger can be seen in \cref{fig:68}. Taken together, these results demonstrate the significance of accounting for forces when understanding fine manipulation actions.

\begin{figure}[t!]
    \centering
    \begin{overpic}
        [width=.8\linewidth,trim=0cm 0cm 0cm 0.3cm,clip]{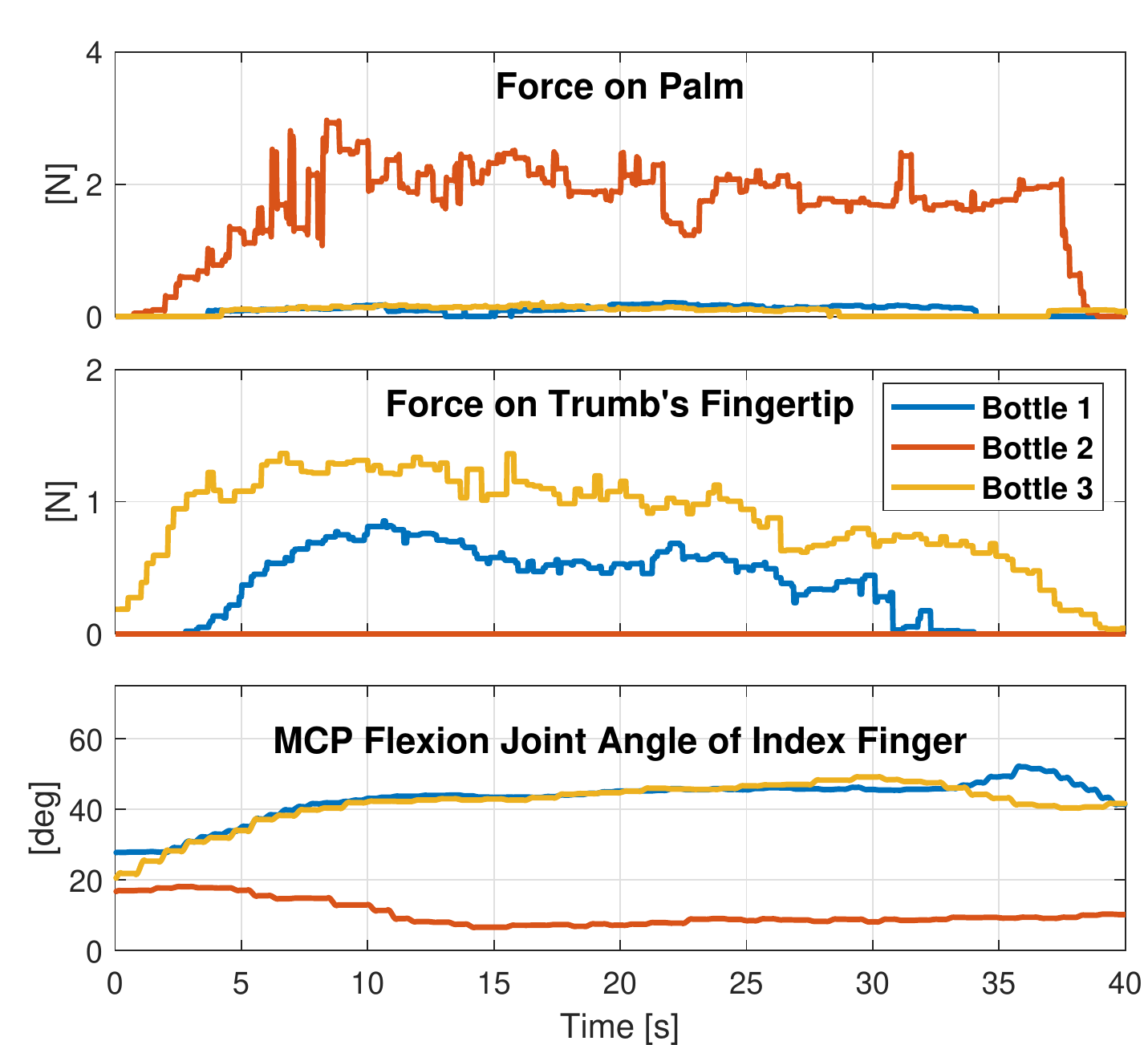}
        \put(37,82.5){\color{black} (a)}
        \put(27,54.5){\color{black} (b)}
        \put(17.5,25.5){\color{black} (c)}
    \end{overpic}
    \caption{\textbf{Force and joint angle recorded by the tactile-sensing glove.} (a) The forces exerted by the palm, (b) forces exerted by the thumb's fingertip, and (c) the flexion angle of the index finger's MCP joint can disentangle the grasp actions of opening different bottles. Reproduced from Ref.~\cite{liu2017glove} with permission.}
    \label{fig:app:tactile:compare}
\end{figure}

Quantitatively, \cref{fig:app:tactile:compare} illustrates one taxel's force collected on the palm, the thumb's fingertip, and the flexion angle of the index finger's MCP joint. In combination, these three readings can differentiate among the action sequences of opening the three bottles. More specifically, as opening \textit{Bottle 2} involves a pressing action on the lid, the tactile glove successfully captures the high force response on the palm. In contrast, the force reading in the same region is almost zero when opening the other two bottles. \textit{Bottle 3}'s pinch-to-open lock necessitates a greater force exerted by the thumb. Indeed, the opening actions introduce a high force response at the thumb's fingertip, with a longer duration than the actions involved in opening \textit{Bottle 1} without a safety lock. Without contacting the lid, the thumb yields no force response when opening \textit{Bottle 2}. Since opening both \textit{Bottle 1} and \textit{Bottle 3} involves a similar twist action, the measured flexion angles of the index finger's MCP joint are around $50^{\circ}$ in both of these cases. Since only the palm touches the lid and the fingers remain stretched, a small flexion angle occurs when opening \textit{Bottle 2}.

\begin{figure}[t!]
    \centering
    \begin{subfigure}[b]{\linewidth}
        \includegraphics[width=\linewidth]{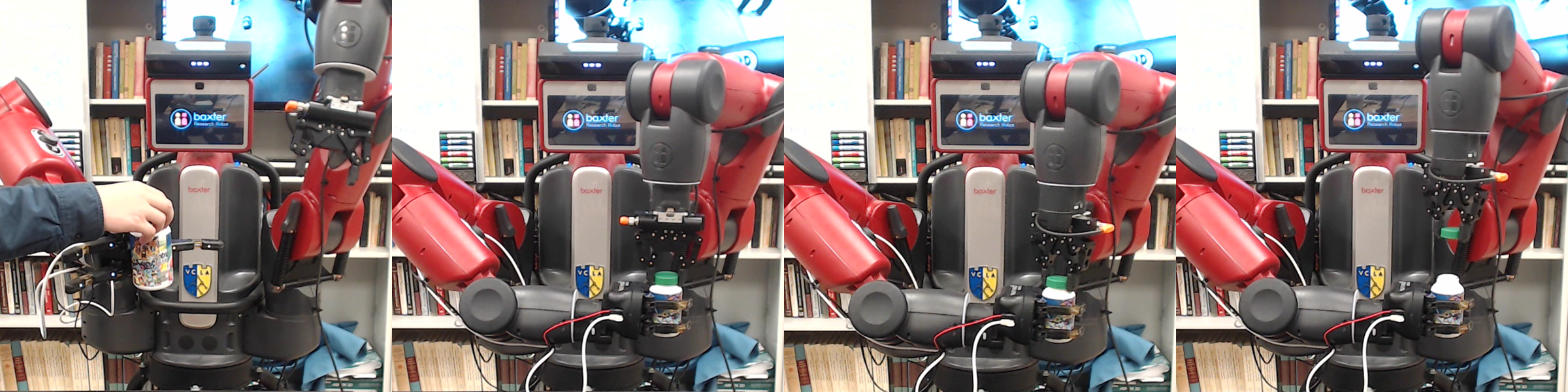}
        \caption{}
        \label{fig:app:tactile:69_exe}
    \end{subfigure}%
    \\
    \begin{subfigure}[b]{\linewidth}
        \includegraphics[width=\linewidth]{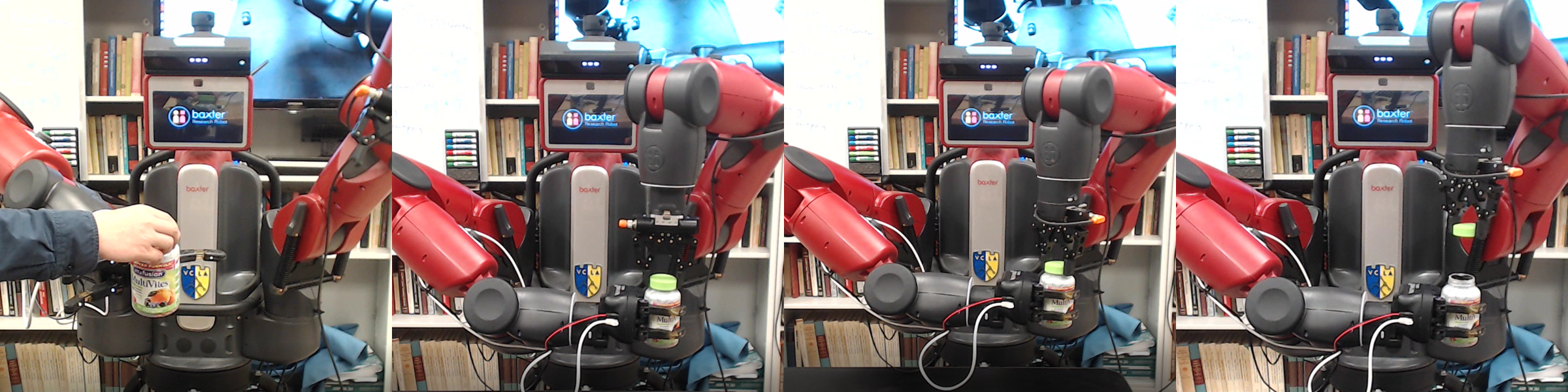}
        \caption{}
        \label{fig:app:tactile:64_exe}
    \end{subfigure}%    
    \caption{\textbf{A Baxter robot learns to open medicine bottles from the collected manipulation data.} Reproduced from Ref.~\cite{edmonds2017seeing} with permission.}
    \label{fig:app:tactile:robot_exec}
\end{figure}

A promising application of the proposed glove is learning fine manipulation actions from human demonstrations. The collected tactile data has facilitated investigations into a robot's functional understanding of actions and imitation learning~\cite{edmonds2017seeing,liu2019mirroring}, inverse reinforcement learning~\cite{xie2019vrgrasp}, and learning explainable models that promote human trust~\cite{edmonds2019tale}. \cref{fig:app:tactile:robot_exec} showcases the robot's learned skills of opening different medicine bottles~\cite{edmonds2017seeing}.

\subsection{VR mode}\label{sec:app:vr}

When operating in VR mode, the reconfigurable glove provides a unique advantage compared with traditional hardware. Below, we showcase two data types that can be collected effectively in this mode.

\paragraph{Trajectories}

\begin{figure}[t!]
    \begin{overpic}
        [width=\linewidth]{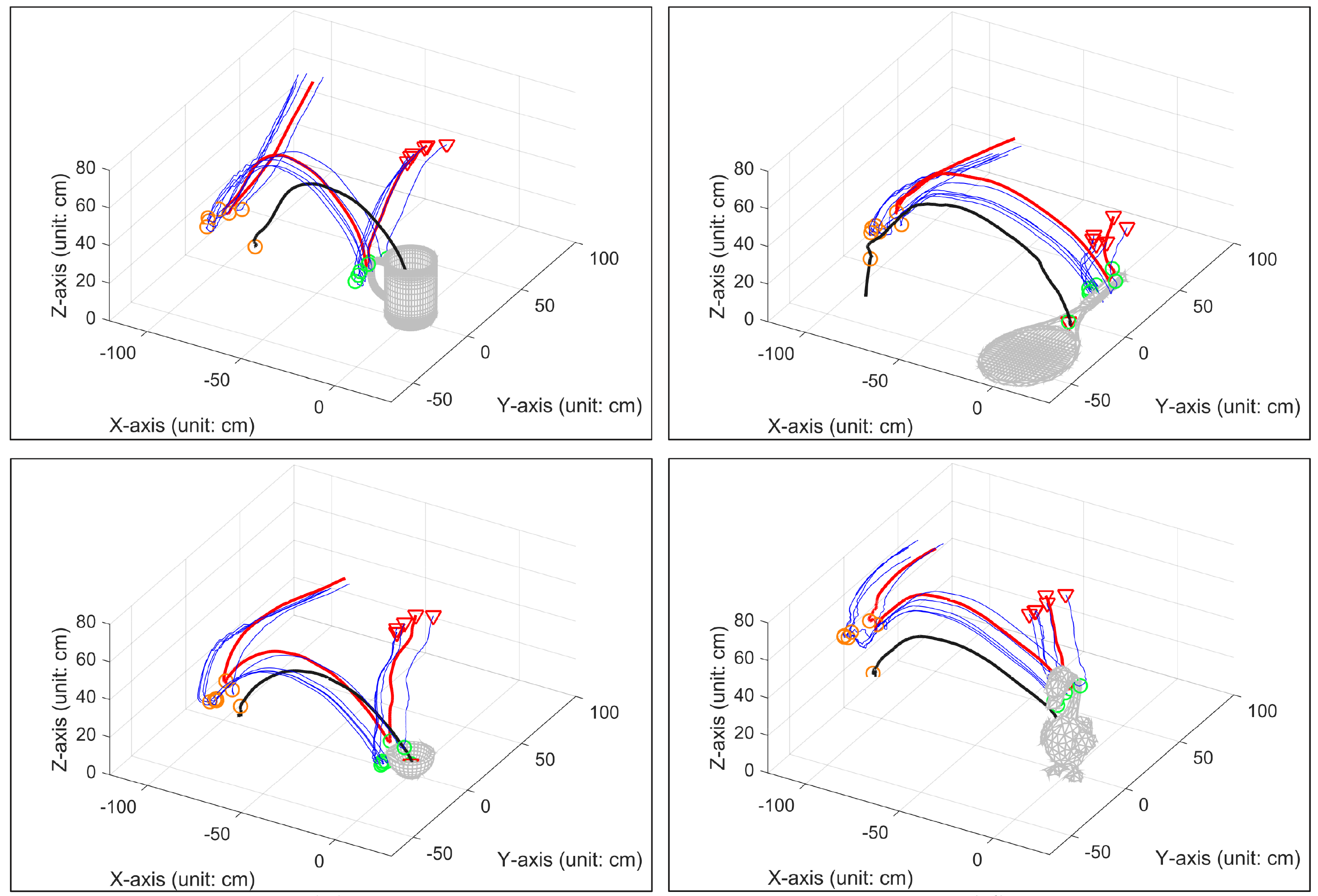}
        \put(23,63.5){\color{black} (a)}
        \put(73,63.5){\color{black} (b)}
        \put(23,29){\color{black} (c)}
        \put(73,29){\color{black} (d)}
    \end{overpic}
    \caption{\textbf{Examples of hand and object trajectories collected by the reconfigurable glove operating in VR mode.} Red triangles indicate the starting poses. The red line and the blue lines show the recorded hand movement and the trajectories of the fingertips, respectively. Once the contact points (green circles) are sufficient to trigger a stable grasp, the object moves together with the hand, following the black line, until the grasp becomes unstable--that is, until it is released at the orange circles. Reproduced from Ref.~\cite{liu2019high} with permission.}
    \label{fig:app:vr:traj}
\end{figure}
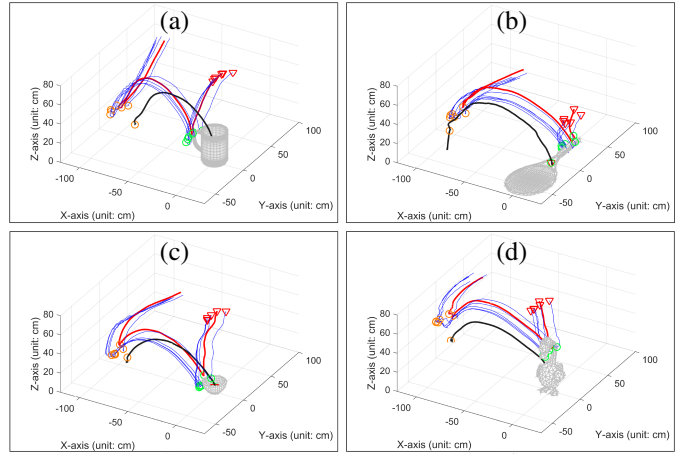

Hand and object trajectories are particularly useful in robot learning from demonstration. Diverse object models can be placed in the VR without setting up a physical apparatus to ensure a natural hand trajectory. \cref{fig:app:vr:traj} shows some qualitative results of collected trajectories: the hand movement (red line) and the five fingertips' trajectories (blue lines) by combining global hand pose and hand gesture sensing, and the grasped object's movement (black line) as the result of hand movement and grasp configuration (stable grasp or not). These results demonstrate the reliability of our design and the richness of the collected trajectory information in a manipulation event.

\paragraph{Contact points}

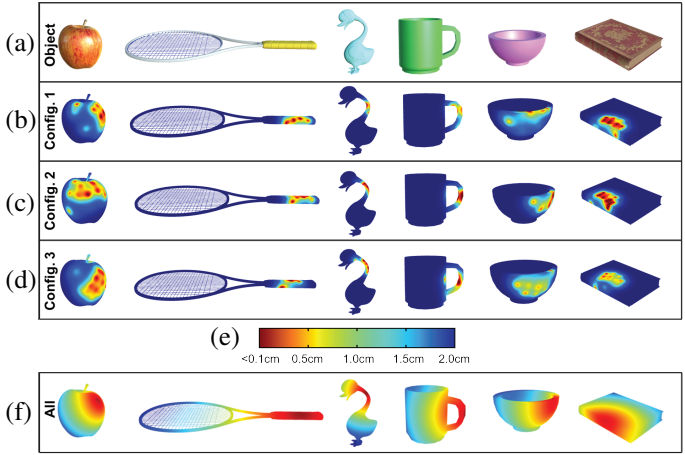
\begin{figure}[t!]
    \centering
    \hfill
    \begin{overpic}
        [width=.96\linewidth]{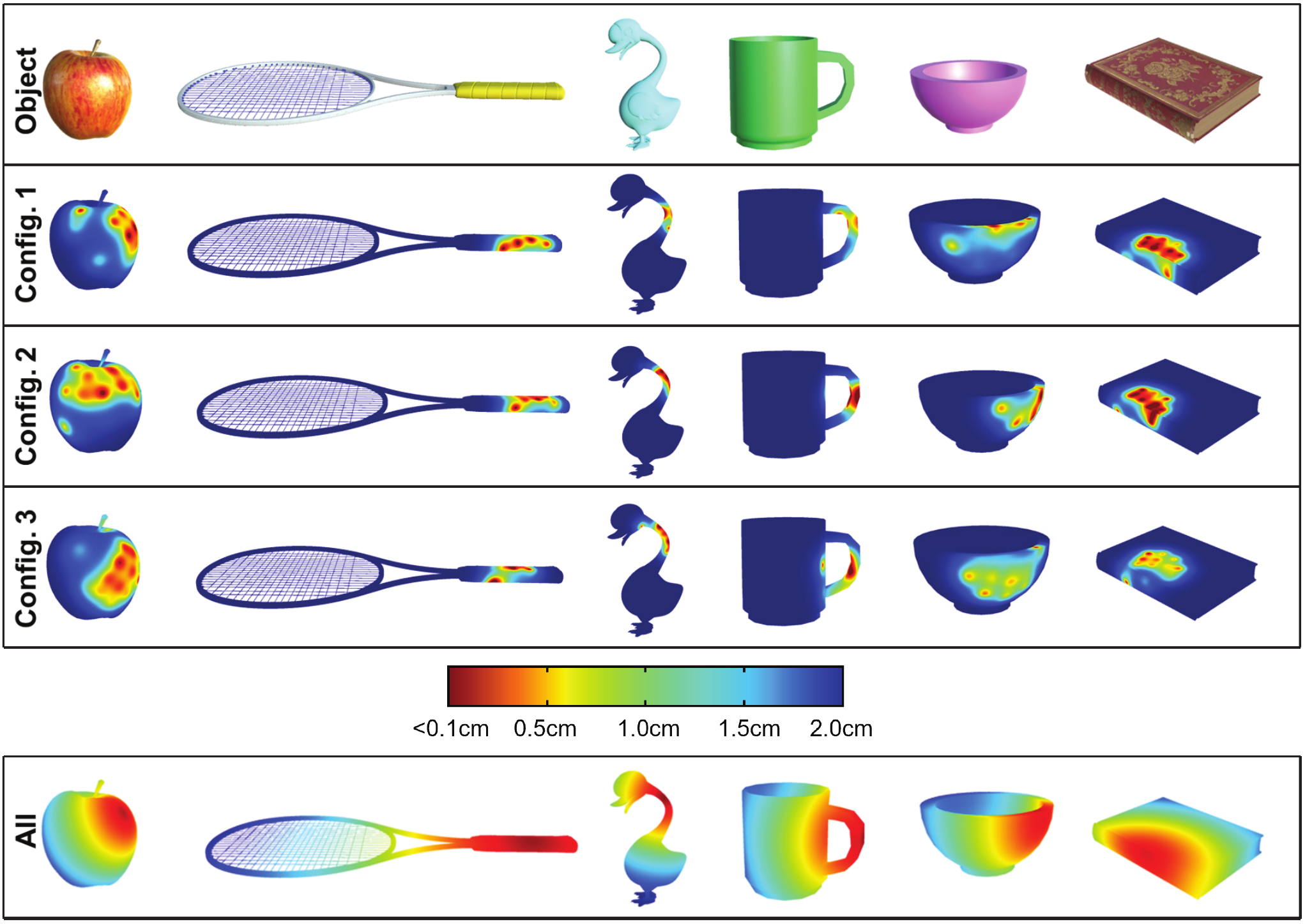}
        \put(-5,63){\color{black} (a)}
        \put(-5,51){\color{black} (b)}
        \put(-5,38){\color{black} (c)}
        \put(-5,26){\color{black} (d)}
        \put(26.7,17.1){\color{black} (e)}
        \put(-5,5.5){\color{black} (f)}
    \end{overpic}
    \caption{\textbf{Contact points in grasping various objects.} (a) Objects to be grasped. (b–d) Three configurations of the contact points performed by different participants. (e) The distance from each contact point. (f) The average of contact points aggregated from all participants, indicating the preferred regions of contact, given the objects.}
    \label{fig:app:vr:contact}
\end{figure}

It is extremely challenging to obtain the contact points of the objects being manipulated. Despite relying As they rely heavily on training data, computer vision-based methods~\cite{rautaray2015vision} are still vulnerable to handling occlusion between hands and objects. Our reconfigurable glove operating in the VR mode can elegantly log this type of data. Given the meshes of the virtual hand model and the object, the VR's physics engine can effectively check the collisions between them. These collisions not only determine whether the object can be stably grasped based on the criteria described in \cref{sec:vr_grasp} but also correspond well to the contact points on the grasped object. By treating a collision point as the spatial center of a spherical volume whose radius is set to the diameter of the finger, \cref{fig:app:vr:contact} shows three configurations of contacts collected from different participants grasping diverse objects. To better uncover the general grasp habits for an object, the contact points shown in the bottom row of \cref{fig:app:vr:contact} are obtained by averaging the spatial positions of contacts across different trails, fitted by a Gaussian distribution.

A fundamental challenge in robot learning of manipulation is the embodiment problem~\cite{dautenhahn2002imitation,liu2019mirroring}: The human hand (five fingers) and robot gripper (usually two or three fingers) have different morphologies. While this problem demands further research, individual contact points can also indicate a preferred region of contact if aggregated from different participants (see the last row in \cref{fig:app:vr:contact}). Such aggregated data can be used for training robot manipulation policies despite different morphologies~\cite{liu2019mirroring}.

\begin{figure*}[t!]
    \centering
    \begin{subfigure}[b]{\linewidth}
        \includegraphics[width=\linewidth]{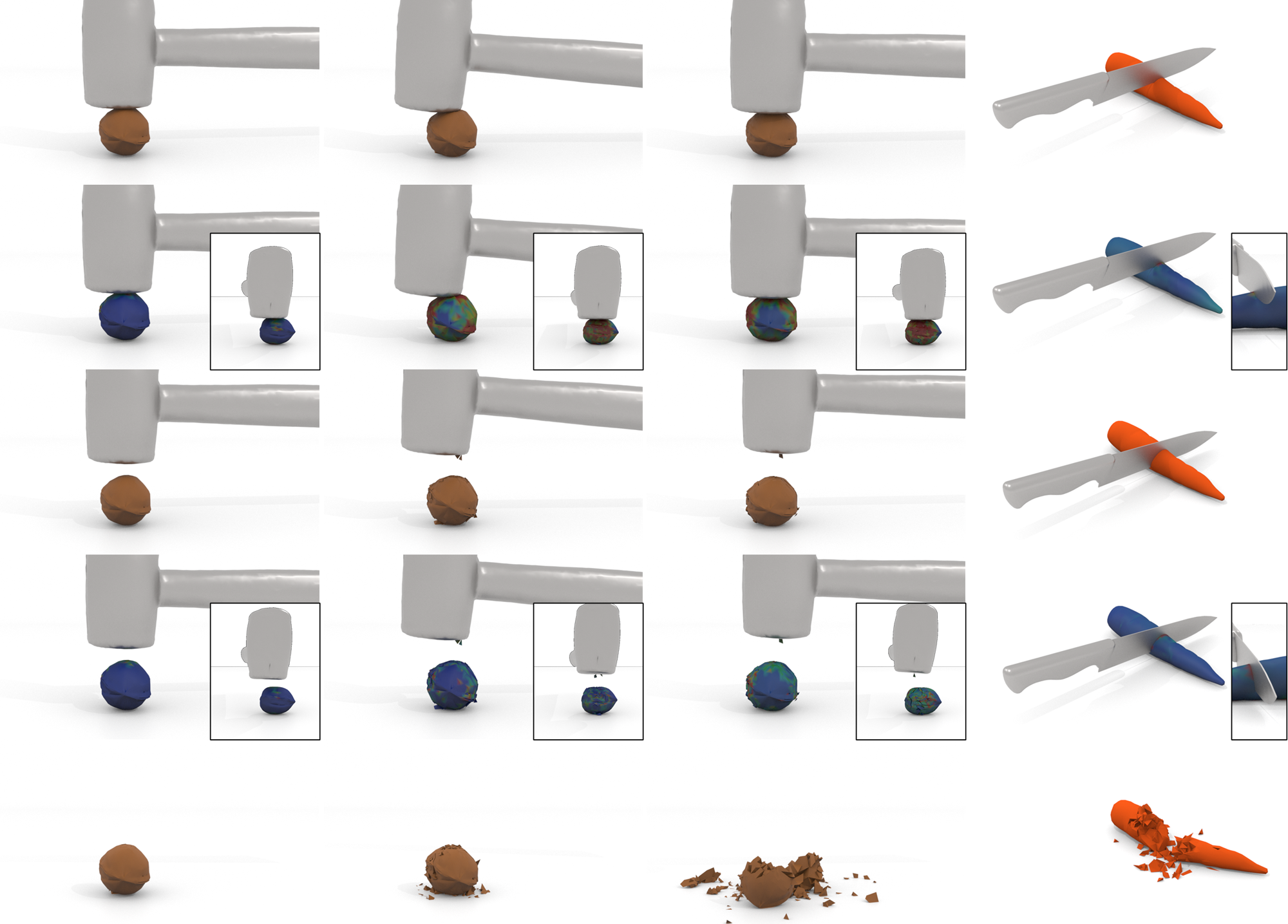}
        \caption{\textbf{Reconstructed tool-use events by simulation.} The first/third rows show the contact moments between the tool and the object. The second/fourth rows are the corresponding stress given by the simulator; red indicates greater stress. The fifth row shows the objects' final status.}
        \label{fig:app:sim:rendering}
    \end{subfigure}%
    \\
    \begin{subfigure}[b]{\linewidth}
        \includegraphics[width=\linewidth]{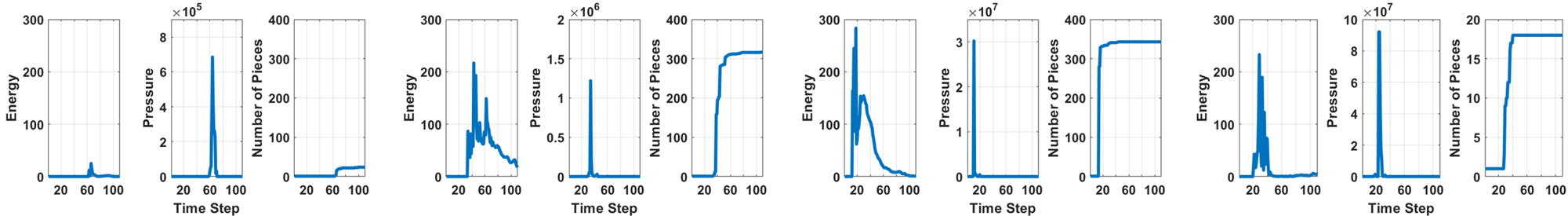}
        \caption{The energy imposed on the objects, the number of fractured pieces, and the contact pressure calculated by the simulator at each time step during tool use.}
        \label{fig:app:sim:data}
    \end{subfigure}%
    \caption{\textbf{Reconstructed 4D manipulation events of tool use by integrating trajectories collected by the reconfigurable glove and physics-based simulation.} This high-fidelity 4D data reveals fine-grained object fluent changes and physical properties at each time step. The results are produced with a simulation at $20~Hz$; one time step is $0.05~s$.}
    \label{fig:app:sim:results}
\end{figure*}

\subsection{Simulation mode}\label{sec:app:sim}

By incorporating the state-of-the-art physics-based simulation, we empower the data glove to capture fine-grained object dynamics during manipulations. \cref{fig:app:sim:results} showcases simulated objects' fluent changes in tool uses. Even when recorded at 120 fps, it is challenging--if not impossible--to capture an object's fluent changes (\eg, how a walnut smashes) using a vision-based method. By feeding the collected trajectory into the simulation, our system renders object fluent changes that are visually similar to the physical reality (see \cref{fig:app:sim:rendering}), thereby revealing critical physical information (see \cref{fig:app:sim:data}) on what occurs in the process.

\paragraph{Results}

\cref{fig:app:sim:rendering} depicts various processes of hammering a walnut. The first column illustrates that a gentle swing action only introduces a small force/energy to the walnut, resulting in a light stress distribution that is quickly eliminated; as a result, the walnut remains uncracked. When a strong swing is performed (third column in \cref{fig:app:sim:rendering}), the larger internal stress causes the walnut to fracture into many pieces, similar to a smashing event in the physical world. This difference is reflected in \cref{fig:app:sim:data}, which was obtained using the physics-based simulator. It is notable that these physical quantities are challenging to measure in the physical world, even with specialized equipment.

\paragraph{Failure examples}

The fourth column of \cref{fig:app:sim:rendering} shows an example of cutting a carrot. The imposed stress is concentrated along the blade that splits the carrot in half. However, when the cutting action is completed and the knife is lifted, it can be seen that the collision between the blade and the carrot has caused undesired fracturing around the cut, which illustrates the limit of the current simulator.

\section{Discussions}

We now discuss two topics in greater depth: Are simulated results good enough, and how do the simulated results help?

\subsection{Are simulated results good enough?}

A central question regarding simulations is whether the simulated results are helpful, given that they are not numerically identical to those directly measured in the physical world. We argue that simulators are indeed helpful, as a simulation preserves the physical events qualitatively, making it possible to study complex events. As illustrated in \cref{fig:app:sim:data}, the walnut's effects have a clear correspondence to the pressure imposed on the contact. Conversely, although a similar amount of energy is imposed when cracking the walnut with a hammer and cutting the carrot with a knife (see the second and fourth columns of hammer and cutting the carrot with a knife (see the second and fourth columns of \cref{fig:app:sim:results}), the resulting pressures differ in magnitude, as the knife introduces a much smaller contact area than the hammer does, producing distinct deformations and topology changes. Hence, the simulation provides a qualitative measurement of the physical events and the objects' fluent change rather than precise quantities. Similar arguments are found in the intuitive physics literature in psychology: Humans usually only make approximate predictions about how states evolve, sometimes even with violations of actual physical laws~\cite{kubricht2017intuitive}. Such inaccuracy does not prevent humans from possessing an effective object and scene understanding; on the contrary, it is a core component of human commonsense knowledge~\cite{spelke2022babies,spelke2007core,zhu2020dark}. Recent work in robot tool use~\cite{zhang2022understanding,li2022gendexgrasp,zhu2015understanding} and physics-informed scene understanding~\cite{han2022scene,han2021reconstructing,chen2019holistic++,huang2018cooperative,huang2018holistic,li2017earthquake,zhu2016inferring,zheng2015scene,zheng2013beyond} has also demonstrated the essential role of physics in understanding objects and scenes.

\subsection{How do the simulated results help?}

The fine-grained object effects produced by the simulation open up new venues for studying existing AI and robotics problems. For example, combining task planning and motion planning~\cite{jiao2021efficieint,jiao2021consolidating,jiao2022planning} is a grand challenge in the field of planning. Simulation could help with this challenge in two aspects~\cite{zhang2022understanding}: (i) by grounding ambiguous task symbols to desired outcomes (\eg, the action symbol of ``crack''), and (ii) by modeling implicit goal specifications (\eg, the status of ``cracked''). In addition, simulations can be used to augment existing datasets, such as GARB~\cite{taheri2020grab} and GenDexGrasp~\cite{li2022gendexgrasp} in grasping, and HUMANISE~\cite{wang2022humanise}, CHAIRS~\cite{jiang2022chairs}, and LEMMA~\cite{jia2020lemma} in scene understanding with unobservable information. Ultimately, we hope that this type of 4D data empowered by physics-based simulation can shed light on several profound questions in manipulation: What and why an object is chosen (\ie, the physics involved), how to properly operate that object (\ie, its affordance), what effect the actor is trying to achieve (\ie, the actor's task goals), and what happens when the goal is not achieved (\ie, planning and replanning).

\section{Conclusion}\label{sec:conclusion}

In this study, we presented three different configurations of a glove-based system based on a unified backbone design, which differs from most conventional data gloves that only capture hand gestures. Utilizing piezoresistive Velostat material, the glove's tactile-sensing mode can aggregate the hand force information during manipulation events. In VR mode, the sensed hand gestures can be reconstructed into a virtual hand to facilitate hand-object interactions in VR by incorporating a caging-based approach, resulting in stable grasps and providing vibrational haptic feedback. The simulation mode further uses an FEM simulator to produce fine-grained object fluent changes and physical properties based on hand-related movements, resulting in 4D manipulation events.

We evaluated the components of the system, including the IMUs, Velostat force-sensor taxels, and haptic feedback provided by the vibration motors, to demonstrate the capability and efficacy of the proposed design. By (i) capturing spatiotemporal signals of force and gesture, (ii) recording hand trajectories and contact points on objects, and (iii) collecting 4D manipulations in challenging manipulation events (\eg, tool use), we demonstrated that the proposed glove-based system can play a crucial role in robot learning from humans and in facilitating embodied AI-related research.

\section{Acknowledgments}

The authors would like to thank Mr. Matt Millar and Dr. Xu Xie (Meta) for developing earlier versions of the system, Miss Chen Zhen (BIGAI) for making the nice figures, and five anonymous reviews for constructive feedback. This work is supported in part by the National Key R\&D Program of China (2021ZD0150200) and the Beijing Nova Program.

\section{Appendix}

\paragraph{Supplementary data}

Code and video demos are available at \url{https://sites.google.com/view/engr-glove}.

\paragraph{Compliance with ethics guidelines}

Hangxin Liu, Zeyu Zhang, Ziyuan Jiao, Zhenliang Zhang, Minchen Li, Chenfanfu Jiang, Yixin Zhu, and Song-Chun Zhu declare that they have no conflict of interest or financial conflicts to disclose.

\newpage

% {
% \renewcommand*{\bibfont}{\normalfont\small}
% \let\oldthebibliography\thebibliography
% \let\endoldthebibliography\endthebibliography
% \renewenvironment{thebibliography}[1]{
%   \begin{oldthebibliography}{#1}
%     \setlength{\itemsep}{0em}
%     \setlength{\parskip}{0.1em}
% }
% {
%   \end{oldthebibliography}
% }
% \bibliography{reference}
% }

\bibliographystyle{elsarticle-num}

% sorted in order of appearance
% \bibliographystyle{unsrt}
\bibliography{reference}

\begin{thebibliography}{101}
\providecommand{\natexlab}[1]{#1}
\providecommand{\url}[1]{\texttt{#1}}
\providecommand{\urlprefix}{URL }
\expandafter\ifx\csname urlstyle\endcsname\relax
  \providecommand{\doi}[1]{doi:\discretionary{}{}{}#1}\else
  \providecommand{\doi}[1]{doi:\discretionary{}{}{}\begingroup
  \urlstyle{rm}\url{#1}\endgroup}\fi
\providecommand{\bibinfo}[2]{#2}

\bibitem[{rea(1 05)}]{realsense}
\bibinfo{title}{Intel Realsense},
  \urlprefix\url{https://www.intelrealsense.com/}, \bibinfo{year}{Accessed:
  2023-01-05}.

\bibitem[{lea(1 05)}]{leapmotion}
\bibinfo{title}{Leap Motion Controller},
  \urlprefix\url{https://www.ultraleap.com/tracking/}, \bibinfo{year}{Accessed:
  2023-01-05}.

\bibitem[{Abbeel and Ng(2004)}]{abbeel2004apprenticeship}
\bibinfo{author}{P.~Abbeel}, \bibinfo{author}{A.~Y. Ng},
  \bibinfo{title}{Apprenticeship learning via inverse reinforcement learning},
  in: \bibinfo{booktitle}{Proceedings of International Conference on Machine
  Learning (ICML)}, \bibinfo{year}{2004}.

\bibitem[{Battaglia et~al.(2016)Battaglia, Bianchi, Altobelli, Grioli,
  Catalano, Serio, Santello, and Bicchi}]{battaglia2016thimblesense}
\bibinfo{author}{E.~Battaglia}, \bibinfo{author}{M.~Bianchi},
  \bibinfo{author}{A.~Altobelli}, \bibinfo{author}{G.~Grioli},
  \bibinfo{author}{M.~G. Catalano}, \bibinfo{author}{A.~Serio},
  \bibinfo{author}{M.~Santello}, \bibinfo{author}{A.~Bicchi},
  \bibinfo{title}{ThimbleSense: a fingertip-wearable tactile sensor for grasp
  analysis}, \bibinfo{journal}{IEEE Transactions on Haptics}
  \bibinfo{volume}{9}~(\bibinfo{number}{1}) (\bibinfo{year}{2016})
  \bibinfo{pages}{121--133}.

\bibitem[{Boulic et~al.(1996)Boulic, Rezzonico, and Thalmann}]{boulic1996multi}
\bibinfo{author}{R.~Boulic}, \bibinfo{author}{S.~Rezzonico},
  \bibinfo{author}{D.~Thalmann}, \bibinfo{title}{Multi-finger manipulation of
  virtual objects}, in: \bibinfo{booktitle}{VRST}, \bibinfo{year}{1996}.

\bibitem[{Bourne(2002)}]{bourne2002food}
\bibinfo{author}{M.~Bourne}, \bibinfo{title}{Food texture and viscosity:
  concept and measurement}, \bibinfo{publisher}{Elsevier},
  \bibinfo{year}{2002}.

\bibitem[{Brahmbhatt et~al.(2019)Brahmbhatt, Ham, Kemp, and
  Hays}]{brahmbhatt2019contactdb}
\bibinfo{author}{S.~Brahmbhatt}, \bibinfo{author}{C.~Ham},
  \bibinfo{author}{C.~C. Kemp}, \bibinfo{author}{J.~Hays},
  \bibinfo{title}{ContactDB: Analyzing and predicting grasp contact via thermal
  imaging}, in: \bibinfo{booktitle}{Proceedings of the IEEE Conference on
  Computer Vision and Pattern Recognition (CVPR)}, \bibinfo{year}{2019}.

\bibitem[{Chen et~al.(2019)Chen, Huang, Yuan, Qi, Zhu, and
  Zhu}]{chen2019holistic++}
\bibinfo{author}{Y.~Chen}, \bibinfo{author}{S.~Huang},
  \bibinfo{author}{T.~Yuan}, \bibinfo{author}{S.~Qi}, \bibinfo{author}{Y.~Zhu},
  \bibinfo{author}{S.-C. Zhu}, \bibinfo{title}{Holistic++ scene understanding:
  Single-view 3d holistic scene parsing and human pose estimation with
  human-object interaction and physical commonsense}, in:
  \bibinfo{booktitle}{Proceedings of International Conference on Computer
  Vision (ICCV)}, \bibinfo{year}{2019}.

\bibitem[{Choi et~al.(2021)Choi, Crump, Duriez, Elmquist, Hager, Han, Hearl,
  Hodgins, Jain, Leve et~al.}]{choi2021use}
\bibinfo{author}{H.~Choi}, \bibinfo{author}{C.~Crump},
  \bibinfo{author}{C.~Duriez}, \bibinfo{author}{A.~Elmquist},
  \bibinfo{author}{G.~Hager}, \bibinfo{author}{D.~Han},
  \bibinfo{author}{F.~Hearl}, \bibinfo{author}{J.~Hodgins},
  \bibinfo{author}{A.~Jain}, \bibinfo{author}{F.~Leve}, et~al.,
  \bibinfo{title}{On the use of simulation in robotics: Opportunities,
  challenges, and suggestions for moving forward},
  \bibinfo{journal}{Proceedings of the National Academy of Sciences (PNAS)}
  \bibinfo{volume}{118}~(\bibinfo{number}{1}).

\bibitem[{Choo et~al.(2021)Choo, Zhao, Jiang, Li, Jiang, and
  Soga}]{choo2021barrier}
\bibinfo{author}{J.~Choo}, \bibinfo{author}{Y.~Zhao},
  \bibinfo{author}{Y.~Jiang}, \bibinfo{author}{M.~Li},
  \bibinfo{author}{C.~Jiang}, \bibinfo{author}{K.~Soga}, \bibinfo{title}{A
  barrier method for frictional contact on embedded interfaces},
  \bibinfo{journal}{arXiv preprint arXiv:2107.05814} .

\bibitem[{Cini et~al.(2019)Cini, Ortenzi, Corke, and
  Controzzi}]{cini2019choice}
\bibinfo{author}{F.~Cini}, \bibinfo{author}{V.~Ortenzi},
  \bibinfo{author}{P.~Corke}, \bibinfo{author}{M.~Controzzi},
  \bibinfo{title}{On the choice of grasp type and location when handing over an
  object}, \bibinfo{journal}{Science Robotics}
  \bibinfo{volume}{4}~(\bibinfo{number}{27}).

\bibitem[{Dautenhahn and Nehaniv(2002)}]{dautenhahn2002imitation}
\bibinfo{author}{K.~Dautenhahn}, \bibinfo{author}{C.~L. Nehaniv},
  \bibinfo{title}{Imitation in Animals and Artifacts}, \bibinfo{publisher}{MIT
  Press Cambridge, MA}, \bibinfo{year}{2002}.

\bibitem[{Dipietro et~al.(2008)Dipietro, Sabatini, and
  Dario}]{dipietro2008survey}
\bibinfo{author}{L.~Dipietro}, \bibinfo{author}{A.~M. Sabatini},
  \bibinfo{author}{P.~Dario}, \bibinfo{title}{A survey of glove-based systems
  and their applications}, \bibinfo{journal}{Transactions on Systems, Man, and
  Cybernetics, Part C (Applications and Reviews)}
  \bibinfo{volume}{38}~(\bibinfo{number}{4}) (\bibinfo{year}{2008})
  \bibinfo{pages}{461--482}.

\bibitem[{Duan et~al.(2012)Duan, Parikh, Crandall, and
  Grauman}]{duan2012discovering}
\bibinfo{author}{K.~Duan}, \bibinfo{author}{D.~Parikh},
  \bibinfo{author}{D.~Crandall}, \bibinfo{author}{K.~Grauman},
  \bibinfo{title}{Discovering localized attributes for fine-grained
  recognition}, in: \bibinfo{booktitle}{Proceedings of the IEEE Conference on
  Computer Vision and Pattern Recognition (CVPR)}, \bibinfo{year}{2012}.

\bibitem[{Edmonds et~al.(2019)Edmonds, Gao, Liu, Xie, Qi, Rothrock, Zhu, Wu,
  Lu, and Zhu}]{edmonds2019tale}
\bibinfo{author}{M.~Edmonds}, \bibinfo{author}{F.~Gao},
  \bibinfo{author}{H.~Liu}, \bibinfo{author}{X.~Xie}, \bibinfo{author}{S.~Qi},
  \bibinfo{author}{B.~Rothrock}, \bibinfo{author}{Y.~Zhu},
  \bibinfo{author}{Y.~N. Wu}, \bibinfo{author}{H.~Lu}, \bibinfo{author}{S.-C.
  Zhu}, \bibinfo{title}{A tale of two explanations: Enhancing human trust by
  explaining robot behavior}, \bibinfo{journal}{Science Robotics}
  \bibinfo{volume}{4}~(\bibinfo{number}{37}).

\bibitem[{Edmonds et~al.(2017)Edmonds, Gao, Xie, Liu, Qi, Zhu, Rothrock, and
  Zhu}]{edmonds2017seeing}
\bibinfo{author}{M.~Edmonds}, \bibinfo{author}{F.~Gao},
  \bibinfo{author}{X.~Xie}, \bibinfo{author}{H.~Liu}, \bibinfo{author}{S.~Qi},
  \bibinfo{author}{Y.~Zhu}, \bibinfo{author}{B.~Rothrock},
  \bibinfo{author}{S.-C. Zhu}, \bibinfo{title}{Feeling the Force: Integrating
  Force and Pose for Fluent Discovery through Imitation Learning to Open
  Medicine Bottles}, in: \bibinfo{booktitle}{Proceedings of International
  Conference on Intelligent Robots and Systems (IROS)}, \bibinfo{year}{2017}.

\bibitem[{Fang et~al.(2021)Fang, Li, Jiang, and Kaufman}]{fang2021guaranteed}
\bibinfo{author}{Y.~Fang}, \bibinfo{author}{M.~Li}, \bibinfo{author}{C.~Jiang},
  \bibinfo{author}{D.~M. Kaufman}, \bibinfo{title}{Guaranteed Globally
  Injective 3D Deformation Processing}, \bibinfo{journal}{ACM Transactions on
  Graphics (TOG)} \bibinfo{volume}{40}~(\bibinfo{number}{4}).

\bibitem[{Feix et~al.(2016)Feix, Romero, Schmiedmayer, Dollar, and
  Kragic}]{feix2016grasp}
\bibinfo{author}{T.~Feix}, \bibinfo{author}{J.~Romero}, \bibinfo{author}{H.-B.
  Schmiedmayer}, \bibinfo{author}{A.~M. Dollar}, \bibinfo{author}{D.~Kragic},
  \bibinfo{title}{The grasp taxonomy of human grasp types},
  \bibinfo{journal}{IEEE Transactions on Human-Machine Systems}
  \bibinfo{volume}{46}~(\bibinfo{number}{1}) (\bibinfo{year}{2016})
  \bibinfo{pages}{66--77}.

\bibitem[{Ferguson et~al.(2021)Ferguson, Li, Schneider, Gil-Ureta, Langlois,
  Jiang, Zorin, Kaufman, and Panozzo}]{ferguson2021intersection}
\bibinfo{author}{Z.~Ferguson}, \bibinfo{author}{M.~Li},
  \bibinfo{author}{T.~Schneider}, \bibinfo{author}{F.~Gil-Ureta},
  \bibinfo{author}{T.~Langlois}, \bibinfo{author}{C.~Jiang},
  \bibinfo{author}{D.~Zorin}, \bibinfo{author}{D.~M. Kaufman},
  \bibinfo{author}{D.~Panozzo}, \bibinfo{title}{Intersection-free Rigid Body
  Dynamics}, \bibinfo{journal}{ACM Transactions on Graphics (TOG)}
  \bibinfo{volume}{40}~(\bibinfo{number}{4}).

\bibitem[{Gu et~al.(2015)Gu, Sheng, Liu, and Ou}]{gu2015fine}
\bibinfo{author}{Y.~Gu}, \bibinfo{author}{W.~Sheng}, \bibinfo{author}{M.~Liu},
  \bibinfo{author}{Y.~Ou}, \bibinfo{title}{Fine manipulative action recognition
  through sensor fusion}, in: \bibinfo{booktitle}{Proceedings of International
  Conference on Intelligent Robots and Systems (IROS)}, \bibinfo{year}{2015}.

\bibitem[{Hammond et~al.(2014)Hammond, Meng{\"u}{\c{c}}, and
  Wood}]{hammond2014toward}
\bibinfo{author}{F.~L. Hammond}, \bibinfo{author}{Y.~Meng{\"u}{\c{c}}},
  \bibinfo{author}{R.~J. Wood}, \bibinfo{title}{Toward a modular soft
  sensor-embedded glove for human hand motion and tactile pressure
  measurement}, in: \bibinfo{booktitle}{Proceedings of International Conference
  on Intelligent Robots and Systems (IROS)}, \bibinfo{year}{2014}.

\bibitem[{Han et~al.(2021)Han, Zhang, Jiao, Xie, Zhu, Zhu, and
  Liu}]{han2021reconstructing}
\bibinfo{author}{M.~Han}, \bibinfo{author}{Z.~Zhang},
  \bibinfo{author}{Z.~Jiao}, \bibinfo{author}{X.~Xie},
  \bibinfo{author}{Y.~Zhu}, \bibinfo{author}{S.-C. Zhu},
  \bibinfo{author}{H.~Liu}, \bibinfo{title}{Reconstructing Interactive 3D Scene
  by Panoptic Mapping and CAD Model Alignments}, in:
  \bibinfo{booktitle}{Proceedings of International Conference on Robotics and
  Automation (ICRA)}, \bibinfo{year}{2021}.

\bibitem[{Han et~al.(2022)Han, Zhang, Jiao, Xie, Zhu, Zhu, and
  Liu}]{han2022scene}
\bibinfo{author}{M.~Han}, \bibinfo{author}{Z.~Zhang},
  \bibinfo{author}{Z.~Jiao}, \bibinfo{author}{X.~Xie},
  \bibinfo{author}{Y.~Zhu}, \bibinfo{author}{S.-C. Zhu},
  \bibinfo{author}{H.~Liu}, \bibinfo{title}{Scene reconstruction with
  functional objects for robot autonomy}, \bibinfo{journal}{International
  Journal of Computer Vision (IJCV)}
  \bibinfo{volume}{130}~(\bibinfo{number}{12}) (\bibinfo{year}{2022})
  \bibinfo{pages}{2940--2961}.

\bibitem[{Hegemann et~al.(2013)Hegemann, Jiang, Schroeder, and
  Teran}]{hegemann2013level}
\bibinfo{author}{J.~Hegemann}, \bibinfo{author}{C.~Jiang},
  \bibinfo{author}{C.~Schroeder}, \bibinfo{author}{J.~M. Teran},
  \bibinfo{title}{A level set method for ductile fracture}, in:
  \bibinfo{booktitle}{Proceedings of ACM SIGGRAPH / Eurographics Symposium on
  Computer Animation (SCA)}, \bibinfo{year}{2013}.

\bibitem[{Heiden et~al.(2021)Heiden, Macklin, Narang, Fox, Garg, and
  Ramos}]{heiden2021disect}
\bibinfo{author}{E.~Heiden}, \bibinfo{author}{M.~Macklin},
  \bibinfo{author}{Y.~Narang}, \bibinfo{author}{D.~Fox},
  \bibinfo{author}{A.~Garg}, \bibinfo{author}{F.~Ramos},
  \bibinfo{title}{DiSECt: A Differentiable Simulation Engine for Autonomous
  Robotic Cutting}, in: \bibinfo{booktitle}{Proceedings of Robotics: Science
  and Systems (RSS)}, \bibinfo{year}{2021}.

\bibitem[{Hu et~al.(2020)Hu, Ding, Peng, Liu, and Wen}]{hu2020flexible}
\bibinfo{author}{B.~Hu}, \bibinfo{author}{T.~Ding}, \bibinfo{author}{Y.~Peng},
  \bibinfo{author}{L.~Liu}, \bibinfo{author}{X.~Wen}, \bibinfo{title}{Flexible
  and attachable inertial measurement unit (IMU)-based motion capture
  instrumentation for the characterization of hand kinematics: A pilot study},
  \bibinfo{journal}{Instrumentation Science \& Technology}
  (\bibinfo{year}{2020}) \bibinfo{pages}{1--21}.

\bibitem[{Hu et~al.(2019)Hu, Liu, Spielberg, Tenenbaum, Freeman, Wu, Rus, and
  Matusik}]{hu2019chainqueen}
\bibinfo{author}{Y.~Hu}, \bibinfo{author}{J.~Liu},
  \bibinfo{author}{A.~Spielberg}, \bibinfo{author}{J.~B. Tenenbaum},
  \bibinfo{author}{W.~T. Freeman}, \bibinfo{author}{J.~Wu},
  \bibinfo{author}{D.~Rus}, \bibinfo{author}{W.~Matusik},
  \bibinfo{title}{ChainQueen: A real-time differentiable physical simulator for
  soft robotics}, in: \bibinfo{booktitle}{Proceedings of International
  Conference on Robotics and Automation (ICRA)}, \bibinfo{year}{2019}.

\bibitem[{Huang et~al.(2018{\natexlab{a}})Huang, Qi, Xiao, Zhu, Wu, and
  Zhu}]{huang2018cooperative}
\bibinfo{author}{S.~Huang}, \bibinfo{author}{S.~Qi}, \bibinfo{author}{Y.~Xiao},
  \bibinfo{author}{Y.~Zhu}, \bibinfo{author}{Y.~N. Wu}, \bibinfo{author}{S.-C.
  Zhu}, \bibinfo{title}{Cooperative Holistic Scene Understanding: Unifying 3D
  Object, Layout and Camera Pose Estimation}, in:
  \bibinfo{booktitle}{Proceedings of Advances in Neural Information Processing
  Systems (NeurIPS)}, \bibinfo{year}{2018}{\natexlab{a}}.

\bibitem[{Huang et~al.(2018{\natexlab{b}})Huang, Qi, Zhu, Xiao, Xu, and
  Zhu}]{huang2018holistic}
\bibinfo{author}{S.~Huang}, \bibinfo{author}{S.~Qi}, \bibinfo{author}{Y.~Zhu},
  \bibinfo{author}{Y.~Xiao}, \bibinfo{author}{Y.~Xu}, \bibinfo{author}{S.-C.
  Zhu}, \bibinfo{title}{Holistic 3D Scene Parsing and Reconstruction from a
  Single RGB Image}, in: \bibinfo{booktitle}{Proceedings of European Conference
  on Computer Vision (ECCV)}, \bibinfo{year}{2018}{\natexlab{b}}.

\bibitem[{Ibarz et~al.(2018)Ibarz, Leike, Pohlen, Irving, Legg, and
  Amodei}]{ibarz2018reward}
\bibinfo{author}{B.~Ibarz}, \bibinfo{author}{J.~Leike},
  \bibinfo{author}{T.~Pohlen}, \bibinfo{author}{G.~Irving},
  \bibinfo{author}{S.~Legg}, \bibinfo{author}{D.~Amodei},
  \bibinfo{title}{Reward learning from human preferences and demonstrations in
  Atari}, in: \bibinfo{booktitle}{Proceedings of Advances in Neural Information
  Processing Systems (NeurIPS)}, \bibinfo{year}{2018}.

\bibitem[{Jeong et~al.(2011)Jeong, Lee, and Kim}]{jeong2011finger}
\bibinfo{author}{E.~Jeong}, \bibinfo{author}{J.~Lee}, \bibinfo{author}{D.~Kim},
  \bibinfo{title}{Finger-gesture recognition glove using velostat (ICCAS
  2011)}, in: \bibinfo{booktitle}{International Conference on Control,
  Automation and Systems (ICCAS)}, \bibinfo{year}{2011}.

\bibitem[{Jia et~al.(2020)Jia, Chen, Huang, Zhu, and Zhu}]{jia2020lemma}
\bibinfo{author}{B.~Jia}, \bibinfo{author}{Y.~Chen},
  \bibinfo{author}{S.~Huang}, \bibinfo{author}{Y.~Zhu}, \bibinfo{author}{S.-C.
  Zhu}, \bibinfo{title}{Lemma: A multi-view dataset for learning multi-agent
  multi-task activities}, in: \bibinfo{booktitle}{Proceedings of European
  Conference on Computer Vision (ECCV)}, \bibinfo{year}{2020}.

\bibitem[{Jiang et~al.(2022)Jiang, Liu, Cao, Cui, Chen, Wang, Zhu, and
  Huang}]{jiang2022chairs}
\bibinfo{author}{N.~Jiang}, \bibinfo{author}{T.~Liu}, \bibinfo{author}{Z.~Cao},
  \bibinfo{author}{J.~Cui}, \bibinfo{author}{Y.~Chen},
  \bibinfo{author}{H.~Wang}, \bibinfo{author}{Y.~Zhu},
  \bibinfo{author}{S.~Huang}, \bibinfo{title}{CHAIRS: Towards Full-Body
  Articulated Human-Object Interaction}, \bibinfo{journal}{arXiv preprint
  arXiv:2212.10621} .

\bibitem[{Jiao et~al.(2022)Jiao, Niu, Zhang, Zhu, Zhu, and
  Liu}]{jiao2022planning}
\bibinfo{author}{Z.~Jiao}, \bibinfo{author}{Y.~Niu},
  \bibinfo{author}{Z.~Zhang}, \bibinfo{author}{S.-C. Zhu},
  \bibinfo{author}{Y.~Zhu}, \bibinfo{author}{H.~Liu}, \bibinfo{title}{Planning
  Sequential Tasks on Contact Graph}, in: \bibinfo{booktitle}{Proceedings of
  International Conference on Intelligent Robots and Systems (IROS)},
  \bibinfo{year}{2022}.

\bibitem[{Jiao et~al.(2021{\natexlab{b}})Jiao, Zeyu, Jiang, Han, Zhu, Zhu, and
  Liu}]{jiao2021consolidating}
\bibinfo{author}{Z.~Jiao}, \bibinfo{author}{Z.~Zeyu},
  \bibinfo{author}{X.~Jiang}, \bibinfo{author}{D.~Han}, \bibinfo{author}{S.-C.
  Zhu}, \bibinfo{author}{Y.~Zhu}, \bibinfo{author}{H.~Liu},
  \bibinfo{title}{Consolidating Kinematic Models to Promote Coordinated Mobile
  Manipulations}, in: \bibinfo{booktitle}{Proceedings of International
  Conference on Intelligent Robots and Systems (IROS)},
  \bibinfo{year}{2021}{\natexlab{b}}.

\bibitem[{Jiao et~al.(2021{\natexlab{a}})Jiao, Zeyu, Wang, Han, Zhu, Zhu, and
  Liu}]{jiao2021efficieint}
\bibinfo{author}{Z.~Jiao}, \bibinfo{author}{Z.~Zeyu},
  \bibinfo{author}{W.~Wang}, \bibinfo{author}{D.~Han}, \bibinfo{author}{S.-C.
  Zhu}, \bibinfo{author}{Y.~Zhu}, \bibinfo{author}{H.~Liu},
  \bibinfo{title}{Efficient Task Planning for Mobile Manipulation: a Virtual
  Kinematic Chain Perspective}, in: \bibinfo{booktitle}{Proceedings of
  International Conference on Intelligent Robots and Systems (IROS)},
  \bibinfo{year}{2021}{\natexlab{a}}.

\bibitem[{Kamel et~al.(2008)Kamel, Sayeed, and Ellis}]{kamel2008glove}
\bibinfo{author}{N.~S. Kamel}, \bibinfo{author}{S.~Sayeed},
  \bibinfo{author}{G.~A. Ellis}, \bibinfo{title}{Glove-based approach to online
  signature verification}, \bibinfo{journal}{IEEE Transactions on Pattern
  Analysis and Machine Intelligence (TPAMI)}
  \bibinfo{volume}{30}~(\bibinfo{number}{6}) (\bibinfo{year}{2008})
  \bibinfo{pages}{1109--1113}.

\bibitem[{Kennedy et~al.(2019)Kennedy, Schmeckpeper, Thakur, Jiang, Kumar, and
  Daniilidis}]{kennedy2019autonomous}
\bibinfo{author}{M.~Kennedy}, \bibinfo{author}{K.~Schmeckpeper},
  \bibinfo{author}{D.~Thakur}, \bibinfo{author}{C.~Jiang},
  \bibinfo{author}{V.~Kumar}, \bibinfo{author}{K.~Daniilidis},
  \bibinfo{title}{Autonomous precision pouring from unknown containers},
  \bibinfo{journal}{IEEE Robotics and Automation Letters (RA-L)}
  \bibinfo{volume}{4}~(\bibinfo{number}{3}) (\bibinfo{year}{2019})
  \bibinfo{pages}{2317--2324}.

\bibitem[{Kiani et~al.(2011)Kiani, Maghsoudi, and
  Minaei}]{kiani2011determination}
\bibinfo{author}{M.~Kiani}, \bibinfo{author}{H.~Maghsoudi},
  \bibinfo{author}{S.~Minaei}, \bibinfo{title}{Determination of poisson's ratio
  and young's modulus of red bean grains}, \bibinfo{journal}{Journal of Food
  Process Engineering} \bibinfo{volume}{34}~(\bibinfo{number}{5})
  (\bibinfo{year}{2011}) \bibinfo{pages}{1573--1583}.

\bibitem[{Kokic et~al.(2017)Kokic, Stork, Haustein, and
  Kragic}]{kokic2017affordance}
\bibinfo{author}{M.~Kokic}, \bibinfo{author}{J.~A. Stork},
  \bibinfo{author}{J.~A. Haustein}, \bibinfo{author}{D.~Kragic},
  \bibinfo{title}{Affordance detection for task-specific grasping using deep
  learning}, in: \bibinfo{booktitle}{International Conference on Humanoid
  Robotics (Humanoids)}, \bibinfo{year}{2017}.

\bibitem[{Kortier et~al.(2015)Kortier, Antonsson, Schepers, Gustafsson, and
  Veltink}]{kortier2015hand}
\bibinfo{author}{H.~G. Kortier}, \bibinfo{author}{J.~Antonsson},
  \bibinfo{author}{H.~M. Schepers}, \bibinfo{author}{F.~Gustafsson},
  \bibinfo{author}{P.~H. Veltink}, \bibinfo{title}{Hand pose estimation by
  fusion of inertial and magnetic sensing aided by a permanent magnet},
  \bibinfo{journal}{Transactions on Neural Systems and Rehabilitation
  Engineering} \bibinfo{volume}{23}~(\bibinfo{number}{5})
  (\bibinfo{year}{2015}) \bibinfo{pages}{796--806}.

\bibitem[{Kortier et~al.(2014)Kortier, Sluiter, Roetenberg, and
  Veltink}]{kortier2014assessment}
\bibinfo{author}{H.~G. Kortier}, \bibinfo{author}{V.~I. Sluiter},
  \bibinfo{author}{D.~Roetenberg}, \bibinfo{author}{P.~H. Veltink},
  \bibinfo{title}{Assessment of hand kinematics using inertial and magnetic
  sensors}, \bibinfo{journal}{Journal of Neuroengineering and Rehabilitation}
  \bibinfo{volume}{11}~(\bibinfo{number}{1}) (\bibinfo{year}{2014})
  \bibinfo{pages}{70}.

\bibitem[{Kramer et~al.(2011)Kramer, Majidi, Sahai, and Wood}]{kramer2011soft}
\bibinfo{author}{R.~K. Kramer}, \bibinfo{author}{C.~Majidi},
  \bibinfo{author}{R.~Sahai}, \bibinfo{author}{R.~J. Wood},
  \bibinfo{title}{Soft curvature sensors for joint angle proprioception}, in:
  \bibinfo{booktitle}{Proceedings of International Conference on Intelligent
  Robots and Systems (IROS)}, \bibinfo{year}{2011}.

\bibitem[{Kubricht et~al.(2017)Kubricht, Holyoak, and
  Lu}]{kubricht2017intuitive}
\bibinfo{author}{J.~R. Kubricht}, \bibinfo{author}{K.~J. Holyoak},
  \bibinfo{author}{H.~Lu}, \bibinfo{title}{Intuitive physics: Current research
  and controversies}, \bibinfo{journal}{Trends in cognitive sciences}
  \bibinfo{volume}{21}~(\bibinfo{number}{10}) (\bibinfo{year}{2017})
  \bibinfo{pages}{749--759}.

\bibitem[{Lan et~al.(2021)Lan, Yang, Kaufman, Yao, Li, and
  Jiang}]{lan2021medial}
\bibinfo{author}{L.~Lan}, \bibinfo{author}{Y.~Yang}, \bibinfo{author}{D.~M.
  Kaufman}, \bibinfo{author}{J.~Yao}, \bibinfo{author}{M.~Li},
  \bibinfo{author}{C.~Jiang}, \bibinfo{title}{Medial {IPC}: Accelerated
  Incremental Potential Contact with Medial Elastics}, \bibinfo{journal}{ACM
  Transactions on Graphics (TOG)} \bibinfo{volume}{40}~(\bibinfo{number}{4}).

\bibitem[{Lee and Shin(2016)}]{lee2016feasibility}
\bibinfo{author}{B.~W. Lee}, \bibinfo{author}{H.~Shin},
  \bibinfo{title}{Feasibility study of sitting posture monitoring based on
  piezoresistive conductive film-based flexible force sensor},
  \bibinfo{journal}{Sensors} \bibinfo{volume}{16}~(\bibinfo{number}{1})
  (\bibinfo{year}{2016}) \bibinfo{pages}{15--16}.

\bibitem[{Li et~al.(2017)Li, Liang, Quigley, Zhao, and Yu}]{li2017earthquake}
\bibinfo{author}{C.~Li}, \bibinfo{author}{W.~Liang},
  \bibinfo{author}{C.~Quigley}, \bibinfo{author}{Y.~Zhao},
  \bibinfo{author}{L.-F. Yu}, \bibinfo{title}{Earthquake safety training
  through virtual drills}, \bibinfo{journal}{Proceedings of IEEE Transactions
  on Visualization \& Computer Graph (TVCG)}
  \bibinfo{volume}{23}~(\bibinfo{number}{4}) (\bibinfo{year}{2017})
  \bibinfo{pages}{1275--1284}.

\bibitem[{Li et~al.(2021{\natexlab{a}})Li, Xia, Mart{\'\i}n-Mart{\'\i}n,
  Lingelbach, Srivastava, Shen, Vainio, Gokmen, Dharan, Jain
  et~al.}]{li2021igibson}
\bibinfo{author}{C.~Li}, \bibinfo{author}{F.~Xia},
  \bibinfo{author}{R.~Mart{\'\i}n-Mart{\'\i}n},
  \bibinfo{author}{M.~Lingelbach}, \bibinfo{author}{S.~Srivastava},
  \bibinfo{author}{B.~Shen}, \bibinfo{author}{K.~Vainio},
  \bibinfo{author}{C.~Gokmen}, \bibinfo{author}{G.~Dharan},
  \bibinfo{author}{T.~Jain}, et~al., \bibinfo{title}{Igibson 2.0:
  Object-centric simulation for robot learning of everyday household tasks},
  in: \bibinfo{booktitle}{Conference on Robot Learning (CoRL)},
  \bibinfo{year}{2021}{\natexlab{a}}.

\bibitem[{Li(2020)}]{li2020robust}
\bibinfo{author}{M.~Li}, \bibinfo{title}{Robust and Accurate Simulation of
  Elastodynamics and Contact}, Ph.D. thesis, \bibinfo{school}{University of
  Pennsylvania}, \bibinfo{year}{2020}.

\bibitem[{Li et~al.(2020)Li, Ferguson, Schneider, Langlois, Zorin, Panozzo,
  Jiang, and Kaufman}]{li2020incremental}
\bibinfo{author}{M.~Li}, \bibinfo{author}{Z.~Ferguson},
  \bibinfo{author}{T.~Schneider}, \bibinfo{author}{T.~Langlois},
  \bibinfo{author}{D.~Zorin}, \bibinfo{author}{D.~Panozzo},
  \bibinfo{author}{C.~Jiang}, \bibinfo{author}{D.~M. Kaufman},
  \bibinfo{title}{Incremental potential contact: Intersection-and
  inversion-free, large-deformation dynamics}, \bibinfo{journal}{ACM
  Transactions on Graphics (TOG)} .

\bibitem[{Li et~al.(2019)Li, Gao, Langlois, Jiang, and
  Kaufman}]{li2019decomposed}
\bibinfo{author}{M.~Li}, \bibinfo{author}{M.~Gao},
  \bibinfo{author}{T.~Langlois}, \bibinfo{author}{C.~Jiang},
  \bibinfo{author}{D.~M. Kaufman}, \bibinfo{title}{Decomposed optimization time
  integrator for large-step elastodynamics}, \bibinfo{journal}{ACM Transactions
  on Graphics (TOG)} \bibinfo{volume}{38}~(\bibinfo{number}{4}).

\bibitem[{Li et~al.(2021{\natexlab{b}})Li, Kaufman, and
  Jiang}]{li2020codimensional}
\bibinfo{author}{M.~Li}, \bibinfo{author}{D.~M. Kaufman},
  \bibinfo{author}{C.~Jiang}, \bibinfo{title}{Codimensional Incremental
  Potential Contact}, \bibinfo{journal}{ACM Transactions on Graphics (TOG)}
  \bibinfo{volume}{40}~(\bibinfo{number}{4}).

\bibitem[{Li et~al.(2022)Li, Liu, Li, Zhu, Yang, and Huang}]{li2022gendexgrasp}
\bibinfo{author}{P.~Li}, \bibinfo{author}{T.~Liu}, \bibinfo{author}{Y.~Li},
  \bibinfo{author}{Y.~Zhu}, \bibinfo{author}{Y.~Yang},
  \bibinfo{author}{S.~Huang}, \bibinfo{title}{GenDexGrasp: Generalizable
  Dexterous Grasping}, \bibinfo{journal}{arXiv preprint arXiv:2210.00722} .

\bibitem[{Ligorio and Sabatini(2013)}]{ligorio2013extended}
\bibinfo{author}{G.~Ligorio}, \bibinfo{author}{A.~M. Sabatini},
  \bibinfo{title}{Extended Kalman filter-based methods for pose estimation
  using visual, inertial and magnetic sensors: comparative analysis and
  performance evaluation}, \bibinfo{journal}{Sensors}
  \bibinfo{volume}{13}~(\bibinfo{number}{2}) (\bibinfo{year}{2013})
  \bibinfo{pages}{1919--1941}.

\bibitem[{Lin et~al.(2019)Lin, Lee, and Chen}]{lin2019novel}
\bibinfo{author}{B.-S. Lin}, \bibinfo{author}{I.-J. Lee},
  \bibinfo{author}{J.-L. Chen}, \bibinfo{title}{Novel Assembled Sensorized
  Glove Platform for Comprehensive Hand Function Assessment by Using Inertial
  Sensors and Force Sensing Resistors}, \bibinfo{journal}{Sensors}
  \bibinfo{volume}{20}~(\bibinfo{number}{6}) (\bibinfo{year}{2019})
  \bibinfo{pages}{3379--3389}.

\bibitem[{Lin et~al.(2000)Lin, Wu, and Huang}]{lin2000modeling}
\bibinfo{author}{J.~Lin}, \bibinfo{author}{Y.~Wu}, \bibinfo{author}{T.~S.
  Huang}, \bibinfo{title}{Modeling the constraints of human hand motion}, in:
  \bibinfo{booktitle}{Workshop on Human Motion}, \bibinfo{organization}{IEEE},
  \bibinfo{year}{2000}.

\bibitem[{Liu et~al.(2017{\natexlab{a}})Liu, Xie, Millar, Edmonds, Gao, Zhu,
  Santos, Rothrock, and Zhu}]{liu2017glove}
\bibinfo{author}{H.~Liu}, \bibinfo{author}{X.~Xie},
  \bibinfo{author}{M.~Millar}, \bibinfo{author}{M.~Edmonds},
  \bibinfo{author}{F.~Gao}, \bibinfo{author}{Y.~Zhu}, \bibinfo{author}{V.~J.
  Santos}, \bibinfo{author}{B.~Rothrock}, \bibinfo{author}{S.-C. Zhu},
  \bibinfo{title}{A Glove-based System for Studying Hand-Object Manipulation
  via Joint Pose and Force Sensing}, in: \bibinfo{booktitle}{Proceedings of
  International Conference on Intelligent Robots and Systems (IROS)},
  \bibinfo{year}{2017}{\natexlab{a}}.

\bibitem[{Liu et~al.(2019{\natexlab{a}})Liu, Zhang, Zhu, Jiang, and
  Zhu}]{liu2019mirroring}
\bibinfo{author}{H.~Liu}, \bibinfo{author}{C.~Zhang}, \bibinfo{author}{Y.~Zhu},
  \bibinfo{author}{C.~Jiang}, \bibinfo{author}{S.-C. Zhu},
  \bibinfo{title}{Mirroring without overimitation: Learning functionally
  equivalent manipulation actions}, in: \bibinfo{booktitle}{Proceedings of AAAI
  Conference on Artificial Intelligence (AAAI)},
  \bibinfo{year}{2019}{\natexlab{a}}.

\bibitem[{Liu et~al.(2019{\natexlab{b}})Liu, Zhang, Xie, Zhu, Liu, Wang, and
  Zhu}]{liu2019high}
\bibinfo{author}{H.~Liu}, \bibinfo{author}{Z.~Zhang}, \bibinfo{author}{X.~Xie},
  \bibinfo{author}{Y.~Zhu}, \bibinfo{author}{Y.~Liu},
  \bibinfo{author}{Y.~Wang}, \bibinfo{author}{S.-C. Zhu},
  \bibinfo{title}{High-fidelity grasping in virtual reality using a glove-based
  system}, in: \bibinfo{booktitle}{Proceedings of International Conference on
  Robotics and Automation (ICRA)}, \bibinfo{year}{2019}{\natexlab{b}}.

\bibitem[{Liu et~al.(2022)Liu, Liu, Jiao, Zhu, and Zhu}]{liu2021synthesizing}
\bibinfo{author}{T.~Liu}, \bibinfo{author}{Z.~Liu}, \bibinfo{author}{Z.~Jiao},
  \bibinfo{author}{Y.~Zhu}, \bibinfo{author}{S.-C. Zhu},
  \bibinfo{title}{Synthesizing Diverse and Physically Stable Grasps with
  Arbitrary Hand Structures using Differentiable Force Closure Estimator},
  \bibinfo{journal}{IEEE Robotics and Automation Letters (RA-L)}
  \bibinfo{volume}{7}~(\bibinfo{number}{1}) (\bibinfo{year}{2022})
  \bibinfo{pages}{470--477}.

\bibitem[{Liu et~al.(2017{\natexlab{b}})Liu, Wei, and Zhu}]{liu2017jointly}
\bibinfo{author}{Y.~Liu}, \bibinfo{author}{P.~Wei}, \bibinfo{author}{S.-C.
  Zhu}, \bibinfo{title}{Jointly recognizing object fluents and tasks in
  egocentric videos}, in: \bibinfo{booktitle}{Proceedings of International
  Conference on Computer Vision (ICCV)}, \bibinfo{year}{2017}{\natexlab{b}}.

\bibitem[{Low et~al.(2015)Low, Khin, and Yeow}]{low2015pressure}
\bibinfo{author}{J.~Low}, \bibinfo{author}{P.~Khin}, \bibinfo{author}{C.~Yeow},
  \bibinfo{title}{A pressure-redistributing insole using soft sensors and
  actuators}, in: \bibinfo{booktitle}{Proceedings of International Conference
  on Robotics and Automation (ICRA)}, \bibinfo{year}{2015}.

\bibitem[{Maeda et~al.(2016)Maeda, Ewerton, Koert, and
  Peters}]{maeda2016acquiring}
\bibinfo{author}{G.~Maeda}, \bibinfo{author}{M.~Ewerton},
  \bibinfo{author}{D.~Koert}, \bibinfo{author}{J.~Peters},
  \bibinfo{title}{Acquiring and generalizing the embodiment mapping from human
  observations to robot skills}, \bibinfo{journal}{IEEE Robotics and Automation
  Letters (RA-L)} \bibinfo{volume}{1}~(\bibinfo{number}{2})
  (\bibinfo{year}{2016}) \bibinfo{pages}{784--791}.

\bibitem[{Mahler et~al.(2019)Mahler, Matl, Satish, Danielczuk, DeRose,
  McKinley, and Goldberg}]{mahler2019learning}
\bibinfo{author}{J.~Mahler}, \bibinfo{author}{M.~Matl},
  \bibinfo{author}{V.~Satish}, \bibinfo{author}{M.~Danielczuk},
  \bibinfo{author}{B.~DeRose}, \bibinfo{author}{S.~McKinley},
  \bibinfo{author}{K.~Goldberg}, \bibinfo{title}{Learning ambidextrous robot
  grasping policies}, \bibinfo{journal}{Science Robotics}
  \bibinfo{volume}{4}~(\bibinfo{number}{26}) (\bibinfo{year}{2019})
  \bibinfo{pages}{eaau4984}.

\bibitem[{Mohammadi et~al.(2016)Mohammadi, Baldi, Scheggi, and
  Prattichizzo}]{mohammadi2016fingertip}
\bibinfo{author}{M.~Mohammadi}, \bibinfo{author}{T.~L. Baldi},
  \bibinfo{author}{S.~Scheggi}, \bibinfo{author}{D.~Prattichizzo},
  \bibinfo{title}{Fingertip force estimation via inertial and magnetic sensors
  in deformable object manipulation}, in: \bibinfo{booktitle}{Haptics Symposium
  (HAPTICS)}, \bibinfo{year}{2016}.

\bibitem[{Mohseni-Kabir et~al.(2015)Mohseni-Kabir, Rich, Chernova, Sidner, and
  Miller}]{mohseni2015interactive}
\bibinfo{author}{A.~Mohseni-Kabir}, \bibinfo{author}{C.~Rich},
  \bibinfo{author}{S.~Chernova}, \bibinfo{author}{C.~L. Sidner},
  \bibinfo{author}{D.~Miller}, \bibinfo{title}{Interactive hierarchical task
  learning from a single demonstration}, in: \bibinfo{booktitle}{Proceedings of
  the Tenth Annual ACM/IEEE International Conference on Human-Robot
  Interaction}, \bibinfo{year}{2015}.

\bibitem[{M{\"u}ller et~al.(2015)M{\"u}ller, Schr{\"o}ter, and
  Gross}]{muller2015smart}
\bibinfo{author}{S.~M{\"u}ller}, \bibinfo{author}{C.~Schr{\"o}ter},
  \bibinfo{author}{H.-M. Gross}, \bibinfo{title}{Smart Fur Tactile Sensor for a
  Socially Assistive Mobile Robot}, in: \bibinfo{booktitle}{International
  Conference on Intelligent Robotics and Applications}, \bibinfo{year}{2015}.

\bibitem[{Nagarajan and Grauman(2018)}]{nagarajan2018attributes}
\bibinfo{author}{T.~Nagarajan}, \bibinfo{author}{K.~Grauman},
  \bibinfo{title}{Attributes as operators: factorizing unseen attribute-object
  compositions}, in: \bibinfo{booktitle}{Proceedings of European Conference on
  Computer Vision (ECCV)}, \bibinfo{year}{2018}.

\bibitem[{Newton and Colson(1736)}]{newton1736method}
\bibinfo{author}{I.~Newton}, \bibinfo{author}{J.~Colson}, \bibinfo{title}{The
  Method of Fluxions and Infinite Series; with Its Application to the Geometry
  of Curve-lines}, \bibinfo{publisher}{Henry Woodfall; and sold by John
  Nourse}, \bibinfo{year}{1736}.

\bibitem[{Nguyen et~al.(2016)Nguyen, Kanoulas, Caldwell, and
  Tsagarakis}]{nguyen2016detecting}
\bibinfo{author}{A.~Nguyen}, \bibinfo{author}{D.~Kanoulas},
  \bibinfo{author}{D.~G. Caldwell}, \bibinfo{author}{N.~G. Tsagarakis},
  \bibinfo{title}{Detecting object affordances with convolutional neural
  networks}, in: \bibinfo{booktitle}{Proceedings of International Conference on
  Intelligent Robots and Systems (IROS)}, \bibinfo{year}{2016}.

\bibitem[{Nocedal and Wright(2006)}]{nocedal2006numerical}
\bibinfo{author}{J.~Nocedal}, \bibinfo{author}{S.~Wright},
  \bibinfo{title}{Numerical optimization}, \bibinfo{publisher}{Springer Science
  \& Business Media}, \bibinfo{year}{2006}.

\bibitem[{Oh et~al.(2021)Oh, Kim, Lee, Jeong, Ko, and Bae}]{oh2021liquid}
\bibinfo{author}{J.~Oh}, \bibinfo{author}{S.~Kim}, \bibinfo{author}{S.~Lee},
  \bibinfo{author}{S.~Jeong}, \bibinfo{author}{S.~H. Ko},
  \bibinfo{author}{J.~Bae}, \bibinfo{title}{A liquid metal based multimodal
  sensor and haptic feedback device for thermal and tactile sensation
  generation in virtual reality}, \bibinfo{journal}{Advanced Functional
  Materials} \bibinfo{volume}{31}~(\bibinfo{number}{39}) (\bibinfo{year}{2021})
  \bibinfo{pages}{2007772}.

\bibitem[{Pinto and Gupta(2016)}]{pinto2016supersizing}
\bibinfo{author}{L.~Pinto}, \bibinfo{author}{A.~Gupta},
  \bibinfo{title}{Supersizing self-supervision: Learning to grasp from 50k
  tries and 700 robot hours}, in: \bibinfo{booktitle}{Proceedings of
  International Conference on Robotics and Automation (ICRA)},
  \bibinfo{year}{2016}.

\bibitem[{Prieur et~al.(2012)Prieur, Perdereau, and
  Bernardino}]{prieur2012modeling}
\bibinfo{author}{U.~Prieur}, \bibinfo{author}{V.~Perdereau},
  \bibinfo{author}{A.~Bernardino}, \bibinfo{title}{Modeling and planning
  high-level in-hand manipulation actions from human knowledge and active
  learning from demonstration}, in: \bibinfo{booktitle}{Proceedings of
  International Conference on Intelligent Robots and Systems (IROS)},
  \bibinfo{year}{2012}.

\bibitem[{Pugach et~al.(2016)Pugach, Melnyk, Tolochko, Pitti, and
  Gaussier}]{pugach2016touch}
\bibinfo{author}{G.~Pugach}, \bibinfo{author}{A.~Melnyk},
  \bibinfo{author}{O.~Tolochko}, \bibinfo{author}{A.~Pitti},
  \bibinfo{author}{P.~Gaussier}, \bibinfo{title}{Touch-based admittance control
  of a robotic arm using neural learning of an artificial skin}, in:
  \bibinfo{booktitle}{Proceedings of International Conference on Intelligent
  Robots and Systems (IROS)}, \bibinfo{year}{2016}.

\bibitem[{Rautaray and Agrawal(2015)}]{rautaray2015vision}
\bibinfo{author}{S.~S. Rautaray}, \bibinfo{author}{A.~Agrawal},
  \bibinfo{title}{Vision based hand gesture recognition for human computer
  interaction: a survey}, \bibinfo{journal}{Artificial Intelligence Review}
  \bibinfo{volume}{43}~(\bibinfo{number}{1}) (\bibinfo{year}{2015})
  \bibinfo{pages}{1--54}.

\bibitem[{Santaera et~al.(2015)Santaera, Luberto, Serio, Gabiccini, and
  Bicchi}]{santaera2015low}
\bibinfo{author}{G.~Santaera}, \bibinfo{author}{E.~Luberto},
  \bibinfo{author}{A.~Serio}, \bibinfo{author}{M.~Gabiccini},
  \bibinfo{author}{A.~Bicchi}, \bibinfo{title}{Low-cost, fast and accurate
  reconstruction of robotic and human postures via IMU measurements}, in:
  \bibinfo{booktitle}{Proceedings of International Conference on Robotics and
  Automation (ICRA)}, \bibinfo{year}{2015}.

\bibitem[{Schaal et~al.(2003)Schaal, Ijspeert, and
  Billard}]{schaal2003computational}
\bibinfo{author}{S.~Schaal}, \bibinfo{author}{A.~Ijspeert},
  \bibinfo{author}{A.~Billard}, \bibinfo{title}{Computational approaches to
  motor learning by imitation}, \bibinfo{journal}{Philosophical Transactions of
  the Royal Society of London. Series B: Biological Sciences}
  \bibinfo{volume}{358}~(\bibinfo{number}{1431}) (\bibinfo{year}{2003})
  \bibinfo{pages}{537--547}.

\bibitem[{Spelke(2022)}]{spelke2022babies}
\bibinfo{author}{E.~S. Spelke}, \bibinfo{title}{What Babies Know: Core
  Knowledge and Composition Volume 1}, vol.~\bibinfo{volume}{1},
  \bibinfo{publisher}{Oxford University Press}, \bibinfo{year}{2022}.

\bibitem[{Spelke and Kinzler(2007)}]{spelke2007core}
\bibinfo{author}{E.~S. Spelke}, \bibinfo{author}{K.~D. Kinzler},
  \bibinfo{title}{Core knowledge}, \bibinfo{journal}{Developmental Science}
  \bibinfo{volume}{10}~(\bibinfo{number}{1}) (\bibinfo{year}{2007})
  \bibinfo{pages}{89--96}.

\bibitem[{Szot et~al.(2021)Szot, Clegg, Undersander, Wijmans, Zhao, Turner,
  Maestre, Mukadam, Chaplot, Maksymets, Gokaslan, Vondrus, Dharur, Meier,
  Galuba, Chang, Kira, Koltun, Malik, Savva, and Batra}]{szot2021habitat}
\bibinfo{author}{A.~Szot}, \bibinfo{author}{A.~Clegg},
  \bibinfo{author}{E.~Undersander}, \bibinfo{author}{E.~Wijmans},
  \bibinfo{author}{Y.~Zhao}, \bibinfo{author}{J.~Turner},
  \bibinfo{author}{N.~Maestre}, \bibinfo{author}{M.~Mukadam},
  \bibinfo{author}{D.~Chaplot}, \bibinfo{author}{O.~Maksymets},
  \bibinfo{author}{A.~Gokaslan}, \bibinfo{author}{V.~Vondrus},
  \bibinfo{author}{S.~Dharur}, \bibinfo{author}{F.~Meier},
  \bibinfo{author}{W.~Galuba}, \bibinfo{author}{A.~Chang},
  \bibinfo{author}{Z.~Kira}, \bibinfo{author}{V.~Koltun},
  \bibinfo{author}{J.~Malik}, \bibinfo{author}{M.~Savva},
  \bibinfo{author}{D.~Batra}, \bibinfo{title}{Habitat 2.0: Training Home
  Assistants to Rearrange their Habitat}, in: \bibinfo{booktitle}{Proceedings
  of Advances in Neural Information Processing Systems (NeurIPS)},
  \bibinfo{year}{2021}.

\bibitem[{Taheri et~al.(2020)Taheri, Ghorbani, Black, and
  Tzionas}]{taheri2020grab}
\bibinfo{author}{O.~Taheri}, \bibinfo{author}{N.~Ghorbani},
  \bibinfo{author}{M.~J. Black}, \bibinfo{author}{D.~Tzionas},
  \bibinfo{title}{GRAB: A dataset of whole-body human grasping of objects}, in:
  \bibinfo{booktitle}{Proceedings of European Conference on Computer Vision
  (ECCV)}, \bibinfo{year}{2020}.

\bibitem[{Taylor et~al.(2013)Taylor, Ko, Mastrangelo, and
  Bamberg}]{taylor2013forward}
\bibinfo{author}{T.~Taylor}, \bibinfo{author}{S.~Ko},
  \bibinfo{author}{C.~Mastrangelo}, \bibinfo{author}{S.~J.~M. Bamberg},
  \bibinfo{title}{Forward kinematics using IMU on-body sensor network for
  mobile analysis of human kinematics}, in: \bibinfo{booktitle}{Engineering in
  Medicine and Biology Society (EMBC)}, \bibinfo{organization}{IEEE},
  \bibinfo{year}{2013}.

\bibitem[{Wang et~al.(2020{\natexlab{a}})Wang, Yan, Wang, Cai, Gao, Zeng, Wan,
  Wang, Pan, Yu et~al.}]{wang2020gesture}
\bibinfo{author}{M.~Wang}, \bibinfo{author}{Z.~Yan}, \bibinfo{author}{T.~Wang},
  \bibinfo{author}{P.~Cai}, \bibinfo{author}{S.~Gao},
  \bibinfo{author}{Y.~Zeng}, \bibinfo{author}{C.~Wan},
  \bibinfo{author}{H.~Wang}, \bibinfo{author}{L.~Pan}, \bibinfo{author}{J.~Yu},
  et~al., \bibinfo{title}{Gesture recognition using a bioinspired learning
  architecture that integrates visual data with somatosensory data from
  stretchable sensors}, \bibinfo{journal}{Nature Electronics}
  \bibinfo{volume}{3}~(\bibinfo{number}{9})
  (\bibinfo{year}{2020}{\natexlab{a}}) \bibinfo{pages}{563--570}.

\bibitem[{Wang et~al.(2020{\natexlab{b}})Wang, Li, Fang, Zhang, Gao, Tang,
  Kaufman, and Jiang}]{wang2020hierarchical}
\bibinfo{author}{X.~Wang}, \bibinfo{author}{M.~Li}, \bibinfo{author}{Y.~Fang},
  \bibinfo{author}{X.~Zhang}, \bibinfo{author}{M.~Gao},
  \bibinfo{author}{M.~Tang}, \bibinfo{author}{D.~M. Kaufman},
  \bibinfo{author}{C.~Jiang}, \bibinfo{title}{Hierarchical optimization time
  integration for cfl-rate mpm stepping}, \bibinfo{journal}{ACM Transactions on
  Graphics (TOG)} \bibinfo{volume}{39}~(\bibinfo{number}{3}).

\bibitem[{Wang et~al.(2022)Wang, Chen, Liu, Zhu, Liang, and
  Huang}]{wang2022humanise}
\bibinfo{author}{Z.~Wang}, \bibinfo{author}{Y.~Chen}, \bibinfo{author}{T.~Liu},
  \bibinfo{author}{Y.~Zhu}, \bibinfo{author}{W.~Liang},
  \bibinfo{author}{S.~Huang}, \bibinfo{title}{HUMANISE: Language-conditioned
  Human Motion Generation in 3D Scenes}, in: \bibinfo{booktitle}{Proceedings of
  Advances in Neural Information Processing Systems (NeurIPS)},
  \bibinfo{year}{2022}.

\bibitem[{Wen et~al.(2020)Wen, Sun, He, Shi, Zhu, Zhang, Li, Zhang, and
  Lee}]{wen2020machine}
\bibinfo{author}{F.~Wen}, \bibinfo{author}{Z.~Sun}, \bibinfo{author}{T.~He},
  \bibinfo{author}{Q.~Shi}, \bibinfo{author}{M.~Zhu},
  \bibinfo{author}{Z.~Zhang}, \bibinfo{author}{L.~Li},
  \bibinfo{author}{T.~Zhang}, \bibinfo{author}{C.~Lee}, \bibinfo{title}{Machine
  learning glove using self-powered conductive superhydrophobic triboelectric
  textile for gesture recognition in VR/AR applications},
  \bibinfo{journal}{Advanced Science}
  \bibinfo{volume}{7}~(\bibinfo{number}{14}) (\bibinfo{year}{2020})
  \bibinfo{pages}{2000261}.

\bibitem[{Williams et~al.(2005)Williams, Wright, Truong, Daubert, and
  Vinyard}]{williams2005mechanical}
\bibinfo{author}{S.~H. Williams}, \bibinfo{author}{B.~W. Wright},
  \bibinfo{author}{V.~d. Truong}, \bibinfo{author}{C.~R. Daubert},
  \bibinfo{author}{C.~J. Vinyard}, \bibinfo{title}{Mechanical properties of
  foods used in experimental studies of primate masticatory function},
  \bibinfo{journal}{American Journal of Primatology: Official Journal of the
  American Society of Primatologists}
  \bibinfo{volume}{67}~(\bibinfo{number}{3}) (\bibinfo{year}{2005})
  \bibinfo{pages}{329--346}.

\bibitem[{Wolper et~al.(2019)Wolper, Fang, Li, Lu, Gao, and
  Jiang}]{wolper2019cd}
\bibinfo{author}{J.~Wolper}, \bibinfo{author}{Y.~Fang},
  \bibinfo{author}{M.~Li}, \bibinfo{author}{J.~Lu}, \bibinfo{author}{M.~Gao},
  \bibinfo{author}{C.~Jiang}, \bibinfo{title}{CD-MPM: Continuum damage material
  point methods for dynamic fracture animation}, \bibinfo{journal}{ACM
  Transactions on Graphics (TOG)} \bibinfo{volume}{38}~(\bibinfo{number}{4})
  (\bibinfo{year}{2019}) \bibinfo{pages}{1--15}.

\bibitem[{Xie et~al.(2019{\natexlab{b}})Xie, Li, Zhang, Zhu, and
  Zhu}]{xie2019vrgrasp}
\bibinfo{author}{X.~Xie}, \bibinfo{author}{C.~Li}, \bibinfo{author}{C.~Zhang},
  \bibinfo{author}{Y.~Zhu}, \bibinfo{author}{S.-C. Zhu},
  \bibinfo{title}{Learning Virtual Grasp with Failed Demonstrations via
  Bayesian Inverse Reinforcement Learning}, in: \bibinfo{booktitle}{Proceedings
  of International Conference on Intelligent Robots and Systems (IROS)},
  \bibinfo{year}{2019}{\natexlab{b}}.

\bibitem[{Xie et~al.(2019{\natexlab{a}})Xie, Liu, Zhang, Qiu, Gao, Qi, Zhu, and
  Zhu}]{xie2019vrgym}
\bibinfo{author}{X.~Xie}, \bibinfo{author}{H.~Liu}, \bibinfo{author}{Z.~Zhang},
  \bibinfo{author}{Y.~Qiu}, \bibinfo{author}{F.~Gao}, \bibinfo{author}{S.~Qi},
  \bibinfo{author}{Y.~Zhu}, \bibinfo{author}{S.-C. Zhu}, \bibinfo{title}{Vrgym:
  A virtual testbed for physical and interactive ai}, in:
  \bibinfo{booktitle}{Proceedings of the ACM Turing Celebration
  Conference-China}, \bibinfo{year}{2019}{\natexlab{a}}.

\bibitem[{Xiong et~al.(2016)Xiong, Shukla, Xiong, and Zhu}]{xiong2016robot}
\bibinfo{author}{C.~Xiong}, \bibinfo{author}{N.~Shukla},
  \bibinfo{author}{W.~Xiong}, \bibinfo{author}{S.-C. Zhu},
  \bibinfo{title}{Robot learning with a spatial, temporal, and causal and-or
  graph}, in: \bibinfo{booktitle}{Proceedings of International Conference on
  Robotics and Automation (ICRA)}, \bibinfo{year}{2016}.

\bibitem[{Yahya et~al.(2017)Yahya, Li, Kalakrishnan, Chebotar, and
  Levine}]{yahya2017collective}
\bibinfo{author}{A.~Yahya}, \bibinfo{author}{A.~Li},
  \bibinfo{author}{M.~Kalakrishnan}, \bibinfo{author}{Y.~Chebotar},
  \bibinfo{author}{S.~Levine}, \bibinfo{title}{Collective robot reinforcement
  learning with distributed asynchronous guided policy search}, in:
  \bibinfo{booktitle}{Proceedings of International Conference on Intelligent
  Robots and Systems (IROS)}, \bibinfo{year}{2017}.

\bibitem[{Zeng et~al.(2018)Zeng, Song, Yu, Donlon, Hogan, Bauza, Ma, Taylor,
  Liu, Romo et~al.}]{zeng2018robotic}
\bibinfo{author}{A.~Zeng}, \bibinfo{author}{S.~Song}, \bibinfo{author}{K.-T.
  Yu}, \bibinfo{author}{E.~Donlon}, \bibinfo{author}{F.~R. Hogan},
  \bibinfo{author}{M.~Bauza}, \bibinfo{author}{D.~Ma},
  \bibinfo{author}{O.~Taylor}, \bibinfo{author}{M.~Liu},
  \bibinfo{author}{E.~Romo}, et~al., \bibinfo{title}{Robotic pick-and-place of
  novel objects in clutter with multi-affordance grasping and cross-domain
  image matching}, in: \bibinfo{booktitle}{Proceedings of International
  Conference on Robotics and Automation (ICRA)}, \bibinfo{year}{2018}.

\bibitem[{Zhang et~al.(2022)Zhang, Jiao, Wang, Zhu, Zhu, and
  Liu}]{zhang2022understanding}
\bibinfo{author}{Z.~Zhang}, \bibinfo{author}{Z.~Jiao},
  \bibinfo{author}{W.~Wang}, \bibinfo{author}{Y.~Zhu}, \bibinfo{author}{S.-C.
  Zhu}, \bibinfo{author}{H.~Liu}, \bibinfo{title}{Understanding physical
  effects for effective tool-use}, \bibinfo{journal}{IEEE Robotics and
  Automation Letters (RA-L)} \bibinfo{volume}{7}~(\bibinfo{number}{4}).

\bibitem[{Zheng et~al.(2015)Zheng, Zhao, Yu, Ikeuchi, and Zhu}]{zheng2015scene}
\bibinfo{author}{B.~Zheng}, \bibinfo{author}{Y.~Zhao}, \bibinfo{author}{J.~Yu},
  \bibinfo{author}{K.~Ikeuchi}, \bibinfo{author}{S.-C. Zhu},
  \bibinfo{title}{Scene understanding by reasoning stability and safety},
  \bibinfo{journal}{International Journal of Computer Vision (IJCV)}
  \bibinfo{volume}{112}~(\bibinfo{number}{2}) (\bibinfo{year}{2015})
  \bibinfo{pages}{221--238}.

\bibitem[{Zheng et~al.(2013)Zheng, Zhao, Yu, Ikeuchi, and
  Zhu}]{zheng2013beyond}
\bibinfo{author}{B.~Zheng}, \bibinfo{author}{Y.~Zhao}, \bibinfo{author}{J.~C.
  Yu}, \bibinfo{author}{K.~Ikeuchi}, \bibinfo{author}{S.-C. Zhu},
  \bibinfo{title}{Beyond point clouds: Scene understanding by reasoning
  geometry and physics}, in: \bibinfo{booktitle}{Proceedings of the IEEE
  Conference on Computer Vision and Pattern Recognition (CVPR)},
  \bibinfo{year}{2013}.

\bibitem[{Zhu et~al.(2020)Zhu, Gao, Fan, Huang, Edmonds, Liu, Gao, Zhang, Qi,
  Wu et~al.}]{zhu2020dark}
\bibinfo{author}{Y.~Zhu}, \bibinfo{author}{T.~Gao}, \bibinfo{author}{L.~Fan},
  \bibinfo{author}{S.~Huang}, \bibinfo{author}{M.~Edmonds},
  \bibinfo{author}{H.~Liu}, \bibinfo{author}{F.~Gao},
  \bibinfo{author}{C.~Zhang}, \bibinfo{author}{S.~Qi}, \bibinfo{author}{Y.~N.
  Wu}, et~al., \bibinfo{title}{Dark, beyond deep: A paradigm shift to cognitive
  ai with humanlike common sense}, \bibinfo{journal}{Engineering}
  \bibinfo{volume}{6}~(\bibinfo{number}{3}) (\bibinfo{year}{2020})
  \bibinfo{pages}{310--345}.

\bibitem[{Zhu et~al.(2016)Zhu, Jiang, Zhao, Terzopoulos, and
  Zhu}]{zhu2016inferring}
\bibinfo{author}{Y.~Zhu}, \bibinfo{author}{C.~Jiang},
  \bibinfo{author}{Y.~Zhao}, \bibinfo{author}{D.~Terzopoulos},
  \bibinfo{author}{S.-C. Zhu}, \bibinfo{title}{Inferring forces and learning
  human utilities from videos}, in: \bibinfo{booktitle}{Proceedings of the IEEE
  Conference on Computer Vision and Pattern Recognition (CVPR)},
  \bibinfo{year}{2016}.

\bibitem[{Zhu et~al.(2015)Zhu, Zhao, and Zhu}]{zhu2015understanding}
\bibinfo{author}{Y.~Zhu}, \bibinfo{author}{Y.~Zhao}, \bibinfo{author}{S.-C.
  Zhu}, \bibinfo{title}{Understanding tools: {T}ask-oriented object modeling,
  learning and recognition}, in: \bibinfo{booktitle}{Proceedings of the IEEE
  Conference on Computer Vision and Pattern Recognition (CVPR)},
  \bibinfo{year}{2015}.

\bibitem[{Zienkiewicz and Taylor(2000)}]{zienkiewicz2000finite}
\bibinfo{author}{O.~C. Zienkiewicz}, \bibinfo{author}{R.~L. Taylor},
  \bibinfo{title}{The finite element method, vol. 2},
  \bibinfo{publisher}{Butterworth-Heinemann}, \bibinfo{year}{2000}.

\end{thebibliography}
\end{document}